\documentclass[10pt,twocolumn,letterpaper]{article}

\usepackage{cvpr}              

\usepackage[accsupp]{axessibility} 

\usepackage{multirow}
\usepackage{siunitx}
\usepackage{subcaption}

\definecolor{cvprblue}{rgb}{0.21,0.49,0.74}
\usepackage[pagebackref,breaklinks,colorlinks,allcolors=cvprblue]{hyperref}
\usepackage{placeins}

\definecolor{NiceGreen}{RGB}{2,158,115} 
\definecolor{NiceRed}{RGB}{213,94,0}
\newcommand{\up}[1]{\textcolor{NiceGreen}{\scriptsize$\uparrow$#1}}
\newcommand{\down}[1]{\textcolor{NiceRed}{\scriptsize$\downarrow$#1}}

\def\paperID{37558} 
\def\confName{CVPR}
\def\confYear{2026}

\title{Mechanisms of Object Localization in Vision–Language Models}

\author{Timothy Schaumlöffel\textsuperscript{1,2} \quad
Martina G. Vilas\textsuperscript{1} \quad
Gemma Roig\textsuperscript{1,2}\\
{\small \textsuperscript{1}Goethe University Frankfurt, Germany\quad
\textsuperscript{2}The Hessian Center for AI, Germany} \\
{\small\href{https://github.com/t9s9/vlm-loc-mechanisms}{\url{https://github.com/t9s9/vlm-loc-mechanisms}}}
}

\begin{document}
\maketitle
\begin{abstract}
Visually-grounded language models (VLMs) are highly effective in linking visual and textual information, yet they often struggle with basic classification and localization tasks. While classification mechanisms have been studied more extensively, the processes that support object localization remain poorly understood. In this work, we investigate two representative families, LLaVA-1.5 and InternVL-3.5, using a suite of mechanistic interpretability tools, including token ablations, attention knockout, and causal mediation analysis.
We find that localization is driven by a containerization mechanism in which object-aligned tokens define the spatial extent of the object, while the semantic arrangement of tokens within those boundaries is largely irrelevant to the predicted box. Only a very small set of attention heads mediates the causal effect for both classification and localization, concentrating in early–mid layers for LLaVA and mid–late layers for InternVL. The two tasks share some early processing but ultimately depend on largely distinct specialized heads.
Overall, we provide the first layer- and head-level account of localization in VLMs, revealing narrow computational pathways that can guide future model design and grounding objectives.
\end{abstract}

\section{Introduction}
\label{sec:intro}

Visually-grounded Language Models (VLMs) combine a pre-trained vision encoder with a large language model (LLM), typically refined through vision-language instruction tuning. The visual encoder extracts grid-level features from an image, a multimodal adapter maps them into the language embedding space, and the resulting tokens are processed jointly with text by the LLM. This architecture allows VLMs to link visual and textual inputs and has enabled strong performance on tasks such as visual question answering, captioning, and open-ended reasoning about images \cite{liu2023visual,dai2023instructblip,deitke2025molmo,bai2023qwenvlversatilevisionlanguagemodel}.

\noindent Despite these advances, VLMs continue to struggle with core vision tasks. They often misclassify or fail to accurately localize objects \cite{tong2024eyes,zhang2024why}. While the mechanisms underlying classification have been studied \cite{neo2025towards, zhang2024why}, much less is known about localization and detection. 
Closing this gap is important because most VLMs inherit visual features from CLIP \cite{radford2021learning}, which was trained with global image-text supervision and struggles with the pixel-level precision required for localization and detection \cite{bousselham2024grounding,shao2024explore,zhong2022regionclip}. 
Yet VLMs can still answer queries that require identifying and locating objects, suggesting that these models build spatial structure from weakly grounded visual representations. 
This raises the question of how the mechanisms enabling localization and detection emerge in VLMs. \\\\
\noindent In this paper, we present an initial mechanistic study of object localization in VLMs. We combine token-level ablations, controlled perturbations of visual representations, positional decoding, attention knockout, and causal mediation analysis to probe how information relevant for localization is encoded and transformed inside the model. Our main findings are:
\begin{enumerate}[nosep, leftmargin=*]
    \item \textbf{Grounding through containerization.} Localization information is directly encoded in the visual tokens. The model groups these tokens into \textit{containers} that define object boundaries, largely independent of the spatial arrangement of semantics within the object boundaries.

    \item \textbf{Multi-view integration of spatial and semantic cues.} In architectures with global and local views, the global view carries the dominant spatial signal for localization, while local high-resolution crops primarily refine classification, especially for small objects. The two views provide complementary, rather than redundant, evidence.
    
    \item \textbf{Implicit spatial layout learning.} The LLM infers the two-dimensional structure of the image from the one-dimensional token sequence: residual positional signals at the multimodal projection and strong corner anchors are sufficient for the model to reconstruct approximate row boundaries and a grid-like layout.

    \item \textbf{Sparse, task-critical attention heads.} A very small number of attention heads mediate the causal effect for both classification and localization. In LLaVA models, these heads emerge predominantly in the \emph{early–mid layers}, whereas in InternVL they appear in the \emph{mid–late layers}. Despite partial overlap in early processing, the dominant heads for the two tasks are largely disjoint, yet localization causally depends on classification-critical heads, revealing a sequential mechanism in which object identification precedes spatial grounding.
\end{enumerate}
\section{Method} \label{sec:method}
We start by introducing the model architectures, the dataset, and task definitions used throughout this work.

\subsection{Visually Grounded Language Models} \label{subsec:vlm}
We study vision–language models that follow the ViT → MLP → LLM paradigm, where a visual encoder extracts patch features, an MLP projects them into the language space, and a large language model (LLM) generates the output. 
We choose two representative VLMs that instantiate this paradigm at different levels of architectural complexity: LLaVA-1.5 \cite{liu2023visual}, a simple and interpretable baseline, and InternVL-3.5 \cite{wang2025internvl3}, a state-of-the-art variant incorporating token compression and multi-view processing.

\vspace{-1em}
\paragraph{LLaVA-1.5} employs a CLIP ViT-L/14 \cite{radford2021learning} visual backbone and Vicuna LLM \cite{vicuna2023} connected by a two-layer MLP adapter.
Images are padded to square shape and resized to  $336^2$ px. The backbone outputs $24 \times 24$ patch embeddings, which are directly mapped into the LLM embedding space without spatial aggregation. This one-to-one token mapping makes LLaVA a simple, interpretable baseline for analyzing visual–linguistic alignment. We analyze the two available versions: LLaVA-7B and LLaVA-13B, which use Vicuna-7B/13B \cite{vicuna2023} as the language backbone, respectively.

\vspace{-1em}
\paragraph{InternVL-3.5} uses a custom, contrastively pre-trained InternViT-300M backbone \cite{chen2024internvl} and a Qwen3 LLM \cite{yang2025qwen3technicalreport}, linked by a two-layer MLP adapter. It introduces two key architectural extensions that distinguish it from LLaVA:
(i) \textit{Pixel Shuffle}: Each $2 \times 2$ block of visual tokens from the backbone is merged into a single token before projection into the text space using a learned compression. This reduces the number of tokens by a factor of four while preserving local spatial structure.
(ii) \textit{Dynamic High-Resolution Processing}: Input images are split into a variable number of $448^2$ px tiles that are processed independently by the visual backbone. The number of tiles is chosen dynamically based on the image’s aspect ratio and size, enabling more detailed processing.
In parallel, a globally resized $448^2$ px thumbnail provides coarse context. We refer to the high-resolution tiles as local views and the thumbnail as the global view. All local and global tokens are concatenated and passed to the LLM. After compression, each crop produces $16 \times 16$ visual tokens. In our experiments, we cap the number of local tiles at six to reduce computational cost. We study InternVL-3.5 8B, which uses a Qwen3-8B language model.

\subsection{Dataset}\label{subsec:dataset}
To ensure that our analyses isolate the visual evidence used by VLMs, we construct a carefully curated dataset derived from the COCO validation split~\cite{lin2014microsoft}. This requires correcting annotation inaccuracies and filtering images to remove ambiguous or low-quality cases prior to evaluation.

\vspace{-1em}
\paragraph{Base Dataset and Filtering}
We use the COCO validation split and ensure that none of the evaluated models were trained on these samples. Because the split contains multiple annotation issues (e.g., missing objects and coarse masks), we first apply the semi-automatic correction procedure of~\citet{cocorem}. 
We then apply a small set of quality filters to remove extremely small or dominant objects, low-resolution images, and ambiguous cases with multiple valid targets. A detailed description of the filtering steps is provided in Appendix~\ref{sec:apx:data_filtering}. After filtering, the dataset contains \num{6,403} object annotations across \num{3,560} images.


    
    

\vspace{-1em}
\paragraph{Object-Removed Control Set}
Contextual cues can lead to \textit{hallucinated detections}, where models predict the presence of an object solely from background context. To control for this effect, we construct an auxiliary object-removed variant of the dataset. For each image, the target object is removed and the missing region is inpainted using LaMa~\cite{lama}, which reconstructs background structure with high realism. 
We retain only those image pairs for which a model correctly identifies the object in the original image but fails to do so in the inpainted counterpart. This ensures that subsequent analyses rely on real object evidence rather than contextual correlations. Examples of the inpainted dataset are provided in Appendix~\Cref{fig:apx:object_removal_example}. \\

\noindent Because this procedure results in model-dependent subsets, we take the intersection across all three models, yielding \num{2,248} object annotations across \num{1,720} images as our final probing subset.

\subsection{Task} \label{subsec:task}
We evaluate models on two complementary visual tasks: Classifying object presence in the image and localizing its position by providing the bounding box coordinates. For each task, we design a different prompt for the same image.

\vspace{-1em}
\paragraph{Localization.} The model is prompted to predict the bounding box coordinates of a target object. The predicted bounding boxes are parsed and compared against ground-truth annotations using the intersection-over-union (IoU) metric. Performance is measured as the success rate, defined as the proportion of samples where IoU exceeds thresholds of 0.5, 0.7, and 0.9. The final localization score is obtained by averaging over these three thresholds.

\vspace{-1em}
\paragraph{Classification.} The model is prompted to list all objects present in the image, restricted to COCO's category set. A prediction is counted as correct if the ground-truth class name appears anywhere in the model's response. Performance is reported as the proportion of correctly classified instances. We adopt a list-based formulation over a binary alternative to reduce object hallucinations. \\

\noindent For more details, we refer to Appendix \Cref{sec:apx:prompts}.
\section{Experiments} \label{sec:exp}

\subsection{Visual Information Ablation} \label{sec:abl}
We conduct an ablation study to investigate the contribution of visual input tokens to the performance of the VLMs on the classification and localization tasks.

\vspace{-0.5em}
\paragraph{Method.} We ablate visual information at the LLM input, i.e., after the multimodal projection but before positional encodings and autoregressive processing. To remove image-specific content while preserving domain-consistent embedding statistics, we replace the original visual token embeddings with a global average visual embedding computed once over the ImageNet~\cite{deng2009imagenet} validation set. 

\noindent We evaluate four token selection strategies for ablation:
\begin{enumerate}[leftmargin=*,topsep=0pt,label=\roman*),noitemsep]
    \item \textit{Object Tokens}: We project the object mask onto the image token grid and include all tokens that overlap with it by at least one pixel. To probe for boundary sensitivity and context dependence, we shrink or dilate the mask by 1 or 2 token padding. For InternVL models, this procedure is applied to both the local high-resolution and the global thumbnail views of the object. A visualization and details of the masking procedure are provided in the appendix \Cref{fig:apx:abl_pad_ex}.

    \item \textit{Register Tokens}: Global image features are hypothesized to be encoded in register tokens \cite{darcet2024vision}. We therefore select those tokens whose embedding norms exceed two standard deviations above the mean.
    
    \item \textit{Integrated Gradients}: We identify the image tokens most relevant to the model’s decision by computing Integrated Gradients \cite{sundararajan2017axiomatic} with respect to the correct class logits (for classification) or bounding box coordinates (for localization). Tokens are ranked by their attribution magnitude, and the top-$k$ highest-gradient tokens are selected as the most influential ones.
    
    \item \textit{Random Tokens}: As a control, we randomly select $k$ image tokens, repeat the process with three different seeds, and report the mean and standard deviation.
\end{enumerate}

\vspace{-0.5em}
\paragraph{Results.}
As \Cref{tab:abl} shows, across all three models, both localization and classification tasks rely on the information encoded in object tokens. Ablating these tokens results in a significantly larger performance decline compared to removing an equal number of tokens either randomly or via gradient-based selection.
Localization is more affected than classification: removing object tokens reduces localization performance below 10\% accuracy, while classification still succeeds in 20–30\% of cases.
Positive padding around the object further amplifies the effect, while maintaining the original boundaries through negative padding has minimal influence on performance. These findings indicate that the essential information for both tasks resides within the object boundaries.

\begin{table*}[t]
\setlength{\tabcolsep}{6pt}
\sisetup{
  detect-weight=true,
  detect-family=true,
  table-number-alignment = center,
  table-text-alignment = center
}
\centering
\caption{Performance after token ablation. The baseline corresponds to the model without any token removal and serves as a reference for ablations targeting the object mask (with varying padding), highest-gradient tokens, random tokens, and register tokens. We report both absolute accuracy and the corresponding drop relative to the baseline. The average proportion of removed tokens is indicated as a percentage of all image tokens; for InternVL, if applicable, we report the number of removed tokens for the (global, local) views separately. Similar amounts of removed tokens are highlighted in \textbf{bold}. Standard deviations across random seeds are provided in appendix~\Cref{tab:apx:abl_rand_std}.}
\begin{tabular}{
  lrcc|cc||rcc
}
\toprule
Models: & \multicolumn{3}{c}{\textbf{LLaVA 7B}} 
& \multicolumn{2}{c}{\textbf{LLaVA 13B}} 
&  \multicolumn{3}{c}{\textbf{InternVL3.5 8B}} \\
\cmidrule(lr){1-1} \cmidrule(lr){2-4} \cmidrule(lr){5-6}\cmidrule(lr){7-9}
\textbf{\shortstack{Ablation\\Strategy}} & 
\multicolumn{1}{c}{\shortstack{Token\\(\%)}} &
 Loc. (\%)&\multicolumn{1}{c|}{Cls. (\%)} &
 Loc. (\%)&\multicolumn{1}{c||}{Cls. (\%)} &
\multicolumn{1}{c}{\shortstack{Token\\(\%)}} &
 Loc. (\%) &\multicolumn{1}{c}{Cls. (\%)} \\
\midrule
Baseline               & \textit{0}  & 35.34 & 58.10  & 46.98 & 65.30 & \textit{0} & 72.64 & 83.30   \\\midrule
\textminus{} 2 Padding & \textit{1}  & 34.71 \down{0.6} & 57.78 \down{0.3}  & 47.09 \up{0.1} & 65.21 \down{0.1} & \textit{(1,2)}  & 72.73 \up{0.1} & 83.41 \up{0.1} \\
\textminus{} 1 Padding & \textit{3}  & 31.73 \down{3.6} & 55.29 \down{2.8}  & 43.74 \down{3.2} & 65.48 \up{0.2} & \textit{(4,5)} & 72.35 \down{0.3} & 81.85 \down{1.5} \\
\textbf{Object}        & \textbf{\textit{8}} & \textbf{5.92 \down{29.4}} & \textbf{19.44 \down{38.7}} & \textbf{7.37 \down{39.6}} & \textbf{31.41 \down{33.9}} & \textbf{\textit{(13,10)}} & \textbf{11.27 \down{61.4}} & \textbf{33.19 \down{50.1}} \\
+ 1 Padding            & \textit{14} &  0.73 \down{34.6} & 11.25 \down{46.9}  & 0.59 \down{46.4} & 15.12 \down{50.2} & \textit{(23,14)} &  3.77 \down{68.9} & 23.71 \down{59.6}  \\
+ 2 Padding            & \textit{21} &  0.34 \down{35.0} & 10.59 \down{47.5}  & 0.28 \down{46.7} & 12.54 \down{52.8} & \textit{(34,18)} &  2.02 \down{70.6} & 20.73 \down{62.6}   \\\midrule
\multirow{7}{*}{\shortstack{Integrated\\Gradients}} 
 & \textit{1}  & 31.54 \down{3.8} & 52.67 \down{5.4} & 38.80 \down{8.2} & 60.36 \down{4.9} & \textit{1} & 62.46 \down{10.2} & 81.49 \down{1.8} \\
 & \textit{4}  & 22.14 \down{13.2} & 48.40 \down{9.7} & 23.50 \down{23.5} & 54.72 \down{10.6} & \textit{4} & 43.64 \down{29.0} & 77.49 \down{5.8} \\
 & \textit{8}  & \textbf{15.84 \down{19.5}} & \textbf{43.95 \down{14.2}} & \textbf{13.09 \down{33.9}} & \textbf{49.24 \down{16.1}} & \textit{8} & 30.13 \down{42.5} & 74.69 \down{8.6} \\
 & \textit{16} &  8.75 \down{26.6} & 37.28 \down{20.8} &  5.66 \down{41.3} & 44.80 \down{20.5} & \textit{16} & \textbf{16.21 \down{56.4}} & \textbf{71.00 \down{12.3}} \\
 & \textit{32} &  2.79 \down{32.6} & 29.05 \down{29.1} &  1.39 \down{45.6} & 35.72 \down{29.6} & \textit{32} &  6.44 \down{66.2} & 64.86 \down{18.4} \\
 & \textit{48} &  0.86 \down{34.5} & 32.25 \down{25.9} &  0.37 \down{46.6} & 28.65 \down{36.6} & \textit{48} &  3.23 \down{69.4} & 59.52 \down{23.8} \\\midrule
\multirow{7}{*}{\shortstack{Random\\(3 seeds)}} 
 & \textit{1}  & 35.52 \up{0.2} & 57.98 \down{0.1} & 46.74 \down{0.2} & 64.86 \down{0.4} & \textit{1} & 72.32 \down{0.3} & 83.44 \up{0.1} \\
 & \textit{4}  & 35.57 \up{0.2} & 57.35 \down{0.8} & 45.99 \down{1.0} & 64.68 \down{0.6} & \textit{4} & 72.30 \down{0.3} & 83.33 \up{0.0} \\
 & \textit{8}  & \textbf{35.09 \down{0.2}} & \textbf{56.58 \down{1.5}} & \textbf{45.25 \down{1.7}} & \textbf{63.89 \down{1.4}} & \textit{8} & 71.71 \down{0.9} & 83.10 \down{0.2} \\
 & \textit{16} & 33.71 \down{1.6} & 56.39 \down{1.7} & 43.76 \down{3.2} & 63.92 \down{1.4} & \textit{16} & \textbf{70.46 \down{2.2}} & \textbf{83.02 \down{0.3}} \\
 & \textit{32} & 30.43 \down{4.9} & 55.40 \down{2.7} & 39.88 \down{7.1} & 62.54 \down{2.8} & \textit{32} & 66.88 \down{5.8} & 82.62 \down{0.7} \\
 & \textit{48} & 25.65 \down{9.7} & 54.80 \down{3.3} & 34.44 \down{12.5} & 61.34 \down{4.0} & \textit{48} & 59.03 \down{13.6} & 81.69 \down{1.6} \\\midrule
Register              & \textit{1}  & 35.07 \down{0.3} & 59.48 \up{1.4} & 46.53 \down{0.5} & 64.95 \down{0.4} & \textit{(4,4)} & 72.64 \up{0.0} & 83.41 \up{0.1} \\
\bottomrule
\end{tabular}
\label{tab:abl}
    \vspace{-1em}
\end{table*}

\subsubsection{Object Containerization}
\begin{figure}[ht]
    \centering
    \includegraphics[width=\linewidth]{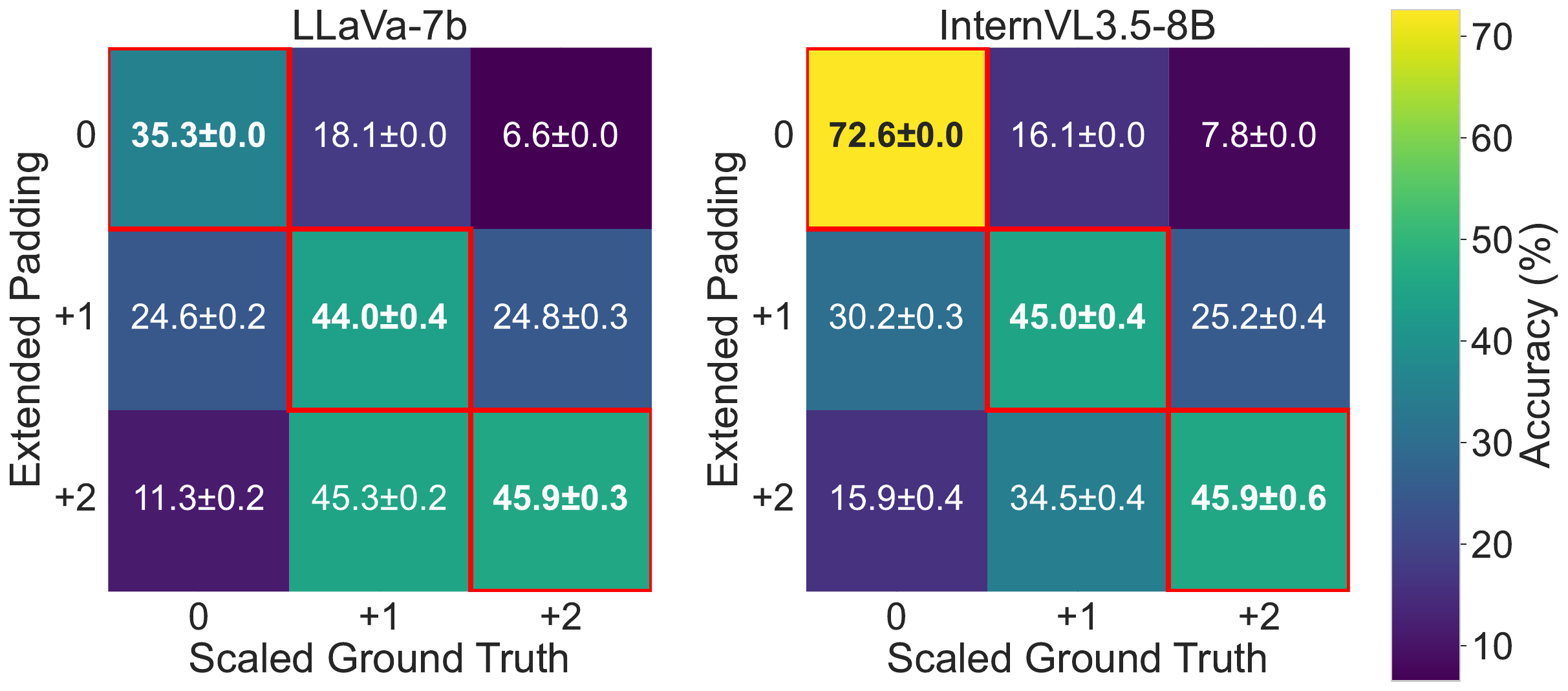}
    \caption{Alignment between predicted and scaled ground-truth bounding boxes under object padding. Each cell shows the mean accuracy between predictions obtained with a given padding level and ground-truth boxes scaled by different amounts. Diagonal entries correspond to matching padding and scaling levels, indicating how well the predicted box size adapts to the artificially enlarged object. Standard deviations are annotated. LLaVA-13B results are shown in the appendix \Cref{fig:apx:ablate_add_object_border}.}
    \label{fig:ablate_add_object_border}
\end{figure}

Next, we investigate the mechanisms by which the model encapsulates objects to generate bounding boxes. To test this, we artificially expand the ground-truth object mask by adding $p$ layers of surrounding tokens. 
Concretely, we randomly duplicate tokens from within the original object and copy them into the adjacent padding region. This procedure increases the spatial extent of the object while disrupting its structure: the added area is filled with misplaced but object-related features (e.g., eye-related tokens may appear below a mouth in a face).
We then measure whether the predicted bounding box expands accordingly across ten random sampling seeds and report the outcome in \Cref{fig:ablate_add_object_border}.  \\

\noindent Across all architectures, the predicted bounding boxes scale consistently with the artificially enlarged objects, as reflected by the strong diagonal alignment: predictions from padding = 1 inputs achieve the highest accuracy with the +1-scaled ground-truth boxes, while padding = 2 inputs best align with the +2-scaled boxes. The results indicate that localization depends mainly on the presence of object-related tokens within the spatial region rather than on their semantically coherent arrangement. We show qualitative examples in appendix \Cref{fig:apx:mask_ext}. \\

\noindent To further support this claim, we shuffle the image tokens within the object mask directly at the LLM input. As shown in \Cref{tab:pertubation}, localization performance drops only slightly under this perturbation compared to a full shuffle of all image tokens. Notably, object classification is unaffected by either shuffling procedure, confirming that the observed effect is specific to localization. \\\\
\noindent Together, these findings suggest that the model employs a form of \textbf{containerization}, in which tokens collectively define the spatial extent of an object, largely independent of their internal semantic structure.

\begin{table}[ht]
    \centering
    \footnotesize
    \label{tab:vlm_perturbations_pm}
    \setlength{\tabcolsep}{3pt}
    \sisetup{
    detect-all,
    table-align-text-post = false,
    table-number-alignment = center,
    separate-uncertainty = true,
    table-space-text-post = \down{99.9}, 
    }
    \caption{Performance under input token shuffling perturbations. We report localization and classification accuracy for two conditions: (i) shuffling all image tokens, and (ii) shuffling image tokens within the object mask. Results are averaged across three seeds.}
    \begin{tabular}{l *{3}{c}}
    \toprule
    Mode & \textbf{LLaVA 7B} & \textbf{LLaVA 13B} & \textbf{InternVL3.5 8B} \\
    \midrule
    \multicolumn{4}{c}{\textit{Localization}} \\
    Baseline & 35.34 & 46.98 & 72.64 \\
    Full &  1.90 $\pm$ 0.10 \down{33.4} &  0.30 $\pm$ 0.00 \down{46.7} &  4.70 $\pm$ 0.20 \down{67.9} \\
    Object & 35.30 $\pm$ 0.20 \down{0.0} & 45.20 $\pm$ 0.20 \down{1.8} & 37.70 $\pm$ 0.40 \down{34.9} \\\midrule
    \multicolumn{4}{c}{\textit{Classification}} \\
     Baseline & 58.10 & 65.30 & 83.30 \\
     Full & 62.42 $\pm$ 0.62 \up{4.3} &  67.04 $\pm$ 0.33 \up{1.7} & 81.19 $\pm$ 0.39 \down{2.1} \\
    Object & 58.44 $\pm$ 0.39 \up{0.3} & 65.21 $\pm$ 0.25 \down{0.1} &  83.15 $\pm$ 0.40 \down{0.1} \\
    \bottomrule
    \end{tabular}
    \label{tab:pertubation}
    \vspace{-1em}
\end{table}

\subsubsection{Contribution of Global and Local Views}
To understand how the InternVL model distributes semantic and spatial information across its two visual pathways, we ablate the object-aligned tokens in either the global resized view or the fine-grained local view. \Cref{tab:abl_local_versus_global} summarizes the resulting change in accuracy for localization and classification. 
Removing object tokens from only one view causes a moderate drop, whereas ablating both views simultaneously leads to a substantially larger decline (see \cref{tab:abl}) for both tasks. This indicates that the model integrates complementary information from the global and local pathways. 
 
\noindent The effect is particularly pronounced for localization: removing the global object tokens reduces accuracy by $-36.4\%$, while local ablation yields a smaller decline of $-9.7\%$. Classification shows the same trend but with reduced magnitude ($-9.5\%$ vs. $-6.6\%$). These results identify the global pathway as the primary carrier of spatial grounding, with local high-resolution cues providing additional semantic detail. 
Padding amplifies this effect. In single-view ablations, increasing the masked region leads to only minor additional degradation because the intact view still provides sufficient object evidence. In two-view ablations, however, padding removes the remaining semantic and spatial cues in both pathways, resulting in much larger drops. This dissociation indicates that each view can compensate for moderate damage to the other, but neither can compensate when both are impaired, demonstrating strong synergy rather than redundancy between the global and local representations.

\begin{table}[h]
\setlength{\tabcolsep}{4pt}
\sisetup{
  detect-weight=true,
  detect-family=true,
  table-number-alignment = center,
  table-text-alignment = center
}
\footnotesize
\centering
\caption{Extended ablation experiment for separate (local and global) views of the \textbf{InternVL} architecture.}
\begin{tabular}{clccc}
\toprule
& \textbf{Strategy} & Token (\%) & Loc. (\%) & \multicolumn{1}{c}{Cls. (\%)} \\
\midrule
& Baseline               & \textit{0}  & 72.64 & 83.30 \\
\midrule
\multirow{3}{*}{\rotatebox[origin=c]{90}{\textit{Local}}}
& \textminus{} 1 Padding & \textit{5}  & 72.41 \down{0.2}  & 82.52 \down{0.8} \\
& \textbf{Object}        & \textbf{\textit{11}}  & \textbf{62.93 \down{9.7}} & \textbf{76.65 \down{6.6}} \\
& + 1 Padding            & \textit{15} & 63.20 \down{9.4} & 77.45 \down{5.9} \\
\midrule
\multirow{3}{*}{\rotatebox[origin=c]{90}{\textit{Global}}}
& \textminus{} 1 Padding & \textit{5}  & 72.58 \down{0.1} & 83.14 \down{0.2} \\
& \textbf{Object}        & \textbf{\textit{14}}  & \textbf{36.20 \down{36.4}} & \textbf{73.80 \down{9.5}} \\
& + 1 Padding            & \textit{24} & 25.86 \down{46.8} & 72.06 \down{11.2} \\
\bottomrule
\end{tabular}
\label{tab:abl_local_versus_global}
\end{table}

\noindent We complement these findings as shown in appendix \Cref{fig:apx:abl_local_versus_global} by breaking down the performance drop by object size.
For single-view ablations, the degradation decreases systematically with object size. Small objects are highly sensitive to the removal of either view, especially for localization (global: -64.8\%; local: -54.3\%). Medium objects show reduced but still substantial dependency. Large objects are comparatively robust: global removal still harms localization (-26.1 \%), whereas local removal can even slightly improve performance (e.g., +6.5 \%), suggesting redundancy or noise in fine-detail tokens for large objects. \\\\
\noindent In summary, we conclude that the global view supplies the essential spatial signal for localization, while the local view primarily supports the classification of small objects. The two pathways therefore provide complementary evidence whose integration enables robust semantic and spatial reasoning. These insights suggest that the number of crops could be adapted dynamically to the task, potentially allowing for more efficient model configurations.

\subsection{Position Encoding} \label{sec:pos_enc}
\begin{figure*}[ht]
    \centering
    \includegraphics[width=\linewidth]{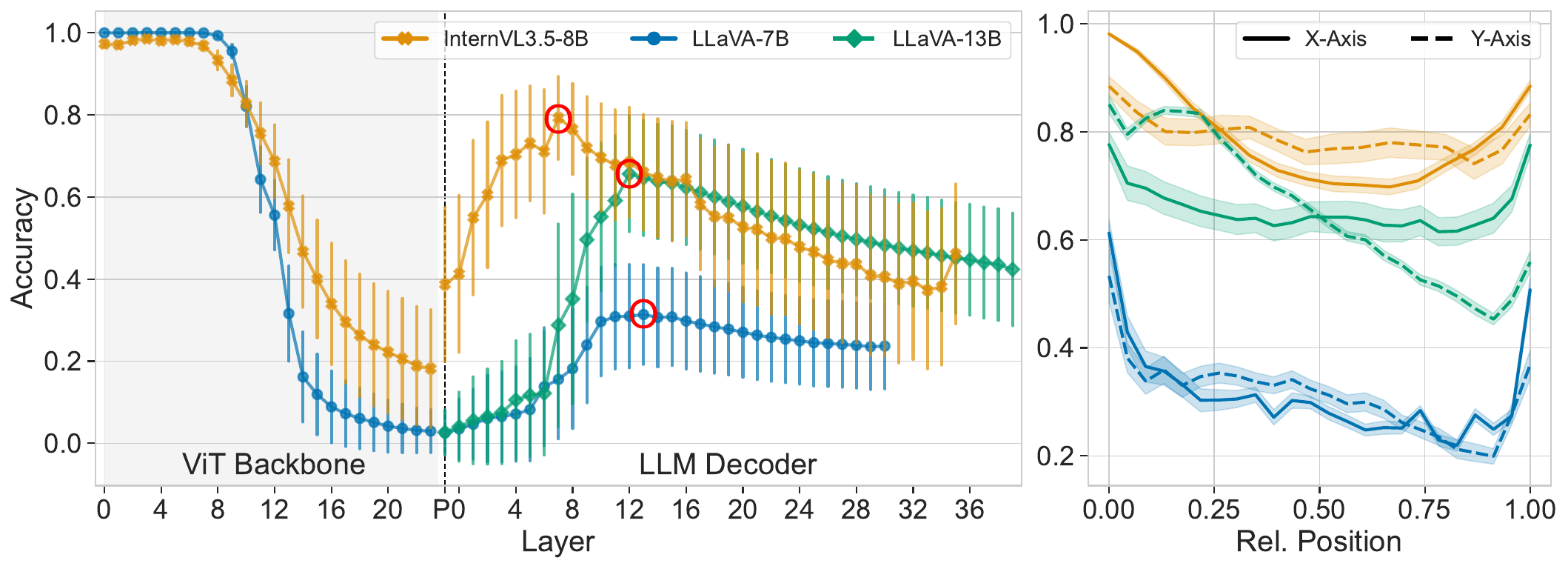}
    \caption{Positional decoding results. Left: average position accuracy per layer for visual backbone (0-23), the multimodal projection (P) and three different language models (0-39). The maximum value per language model is marked with a red circle. Right: per-position accuracy at maximum layer of LLM, showing higher accuracy at the image corners.}
    \label{fig:pos_enc}
    \vspace{-1em}
\end{figure*}

Before examining how object information propagates through the network, we first assess how much of the spatial structure needed for localization is preserved across the model’s layers.

\vspace{-1em}
\paragraph{Method.}
To evaluate how positional information remains identifiable throughout the model’s processing hierarchy, we train a separate linear classifier for each model layer, including the multimodal projection, to predict the position of every image token in the input grid. We predict each spatial axis independently. Classifiers are trained for 10 epochs on 50,000 ImageNet images and evaluated on 10,000 images.

\begin{figure}[ht]
    \centering
    \includegraphics[width=\linewidth]{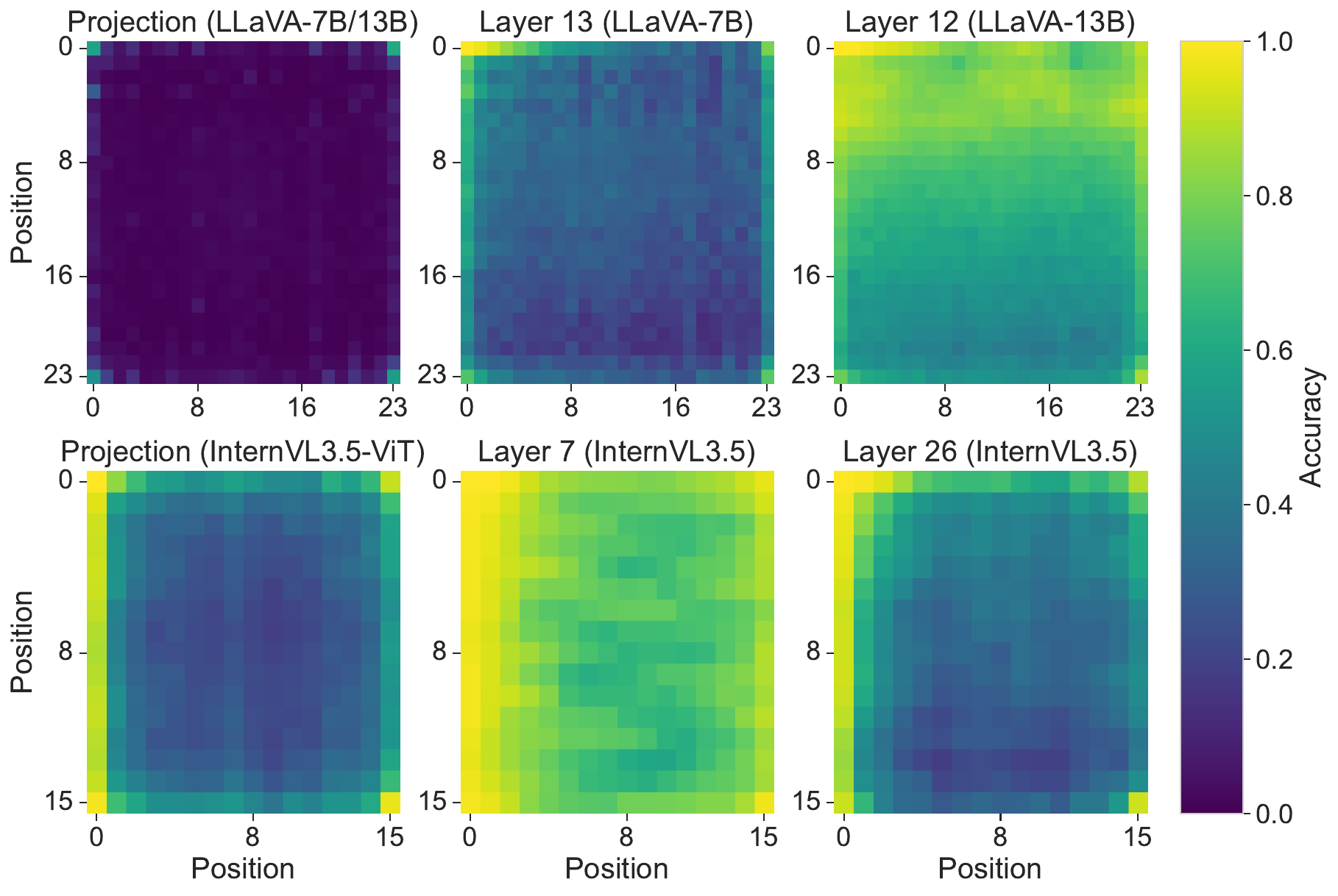}
    \caption{\textbf{Heatmap visualizations of positional decoding accuracy at selected stages of LLaVA.} Each heatmap shows the probability of correctly predicting the position of a visual token in the $24 \times 24$ grid. The multimodal projection retains positional information mainly at the four corners, effectively marking the image boundaries needed to infer its dimensions. Accuracy then increases within the LLM, becoming highest in the early-mid-layers (layer 13 in LLaVA-7B, layer 12 in LLaVA-13B and layer 7 for InternVL3.5). Corners and boundary regions remain the most reliably recovered, indicating that they serve as anchors for reconstructing the global spatial layout.}
    \label{fig:apx:pos_enc_heatmaps}
\end{figure}

\paragraph{Results.} In \Cref{fig:pos_enc}, we observe that positional information in both visual backbones is decodable from early layers but largely vanishes by the final layers, consistent with prior findings that contrastively trained ViTs trade spatial precision for semantic abstraction over depth \cite{jiang2023from}.
In contrast, in the LLM, positional identifiability is initially low. However, it increases rapidly and peaks around layer 12 for LLaVA-7B, layer 13 for LLaVA-13B and layer 7 for InternVL3.5-8B.
The multimodal projection retains strong signals for the four image corners (see app. \cref{fig:apx:pos_enc_heatmaps}), which appear sufficient for the LLM to infer approximate row boundaries (``line breaks'') across the token sequence. Tokens aligned with these inferred line breaks are predicted with higher probability than other positions (\cref{fig:pos_enc}, right), suggesting that the model uses them as structural anchors when reconstructing spatial layouts. 

\subsection{Localizing Task Processing}
Next, we investigate the location of task-specific processing in the language model. Specifically, we examine whether classification and localization rely on shared or distinct components by progressively narrowing the analysis from layer groups to individual attention heads.

\subsubsection{Attention Knockout} \label{sec:attn_knockout}
\begin{figure}[ht]
    \centering
    \includegraphics[width=\linewidth]{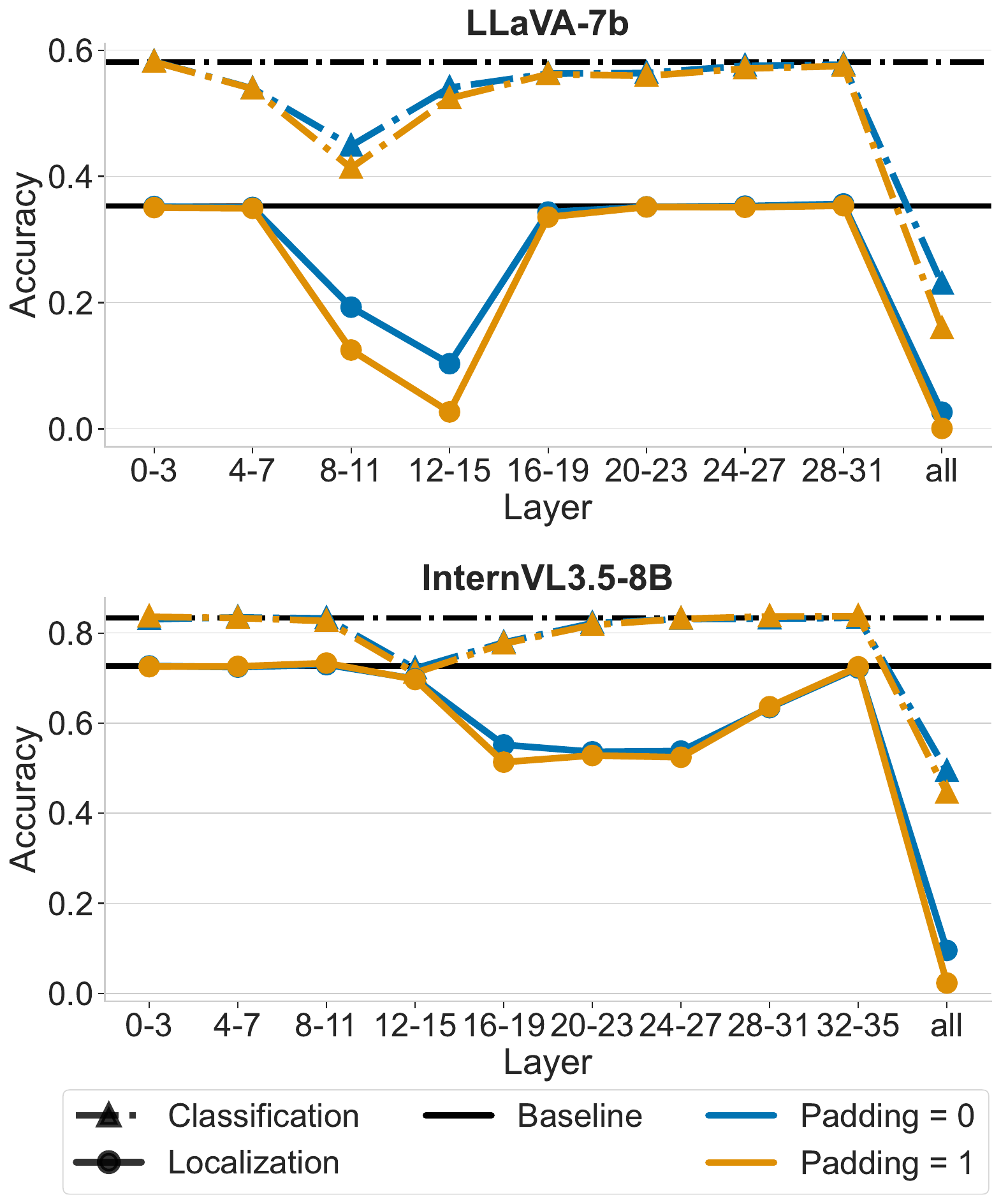}
    \caption{Performance after attention knockout. We block attention from image tokens to object tokens across grouped layers. The curves show classification and localization accuracy and compares to accuracy of the unmodified model. Additional results can be found in appendix \Cref{sec:apx:attn_blk}.}
    \label{fig:attn_block}
    \vspace{-1em}
\end{figure}

\paragraph{Method.} Our next experiments aim to identify where in the network the visual information required for the task is extracted and processed. We apply the \textit{attention knockout} technique \cite{geva-etal-2023-dissecting, neo2025towards}, which blocks attention and thereby prevents communication between tokens. Unlike \cite{neo2025towards}, we eliminate attentions from all tokens following the image tokens to the object tokens, effectively removing any information extracted from the object tokens in those layers. 
We combine layers in groups of four and block all attention heads within each group, as well as across all layers, as a global baseline.
We evaluate the resulting performance drop on classification and localization tasks using our filtered COCO subset. 

\vspace{-0.5em}
\paragraph{Results.} Our results are shown in \Cref{fig:attn_block}. For the LLaVA models, blocking attention to the object significantly decreases performance in the early to mid layers of the model, while perturbing attention flow in later layers leaves performance largely unaffected. In contrast, for InternVL, the strongest decline occurs in the middle to later layers.

\noindent Across all models, both classification and localization rely on early shared layers, after which localization depends on additional task-specific processing. This pattern suggests a two-step processing mechanism: the model first identifies the object, then localizes it. Moreover, for the LLaVA models, the layers with the largest localization decline align with those that retain strong positional information (see \cref{sec:pos_enc}).

\begin{figure}[ht]
    \centering
    \includegraphics[width=\linewidth]{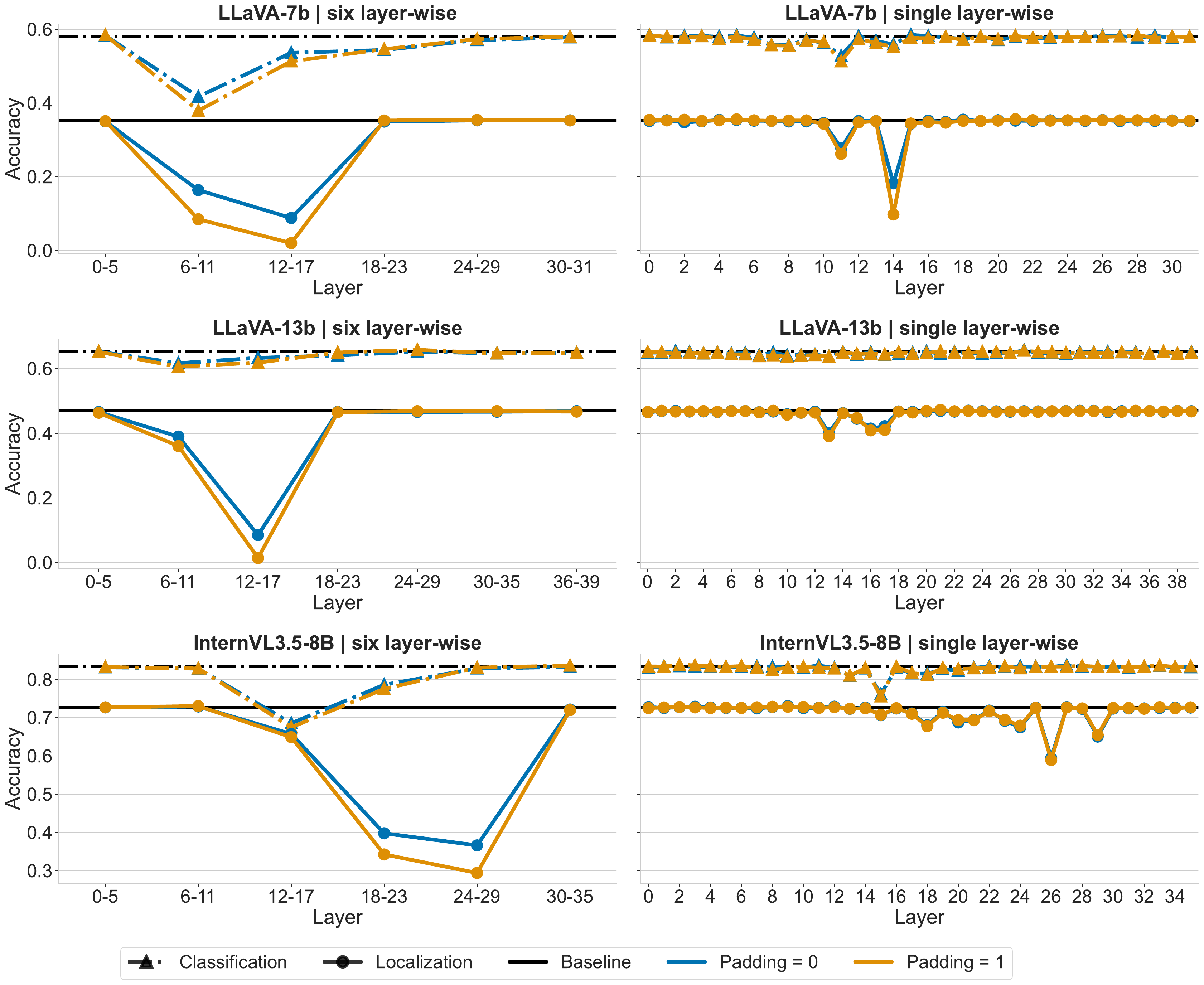}
    \caption{\textbf{Attention blocking across layers.} We measure classification and localization accuracy when blocking attention from post-image tokens to object tokens, either in groups of six layers (left) or one layer at a time (right). Localization accuracy drops sharply in early–mid layers for LLaVA models and in mid–late layers for InternVL, while classification remains largely stable across the network. Blocking attention to padding tokens causes an additional decrease in localization performance.}
    \label{fig:apx:attn_blk_per_layer}
\end{figure}

\noindent Compared to layer-wise knockouts (app. \cref{fig:apx:attn_blk_per_layer} right), ablating groups of layers amplifies the effect, particularly for classification, as further shown in \Cref{fig:apx:attn_blk_per_layer} (left), where we group six consecutive layers.
This suggests that the model accumulates object-related information across several layers rather than relying on layer-specific mechanisms. 
The following section builds on these findings by identifying task-specific heads through causal mediation analysis.

\subsubsection{Causal Mediation Analysis}
In the previous section, we presented evidence that the mechanisms underlying localization and classification are concentrated within a narrow region of the language model’s processing pipeline. Using causal mediation analysis (CMA) \cite{yang2025emergent, wang2023interpretability, Meng2022factual}, we now aim to pinpoint in which model components these mechanisms reside. CMA enables us to estimate the causal contribution of an embedding at a specific computational block within the model.

\begin{figure}[ht]
    \centering
    \includegraphics[width=\linewidth]{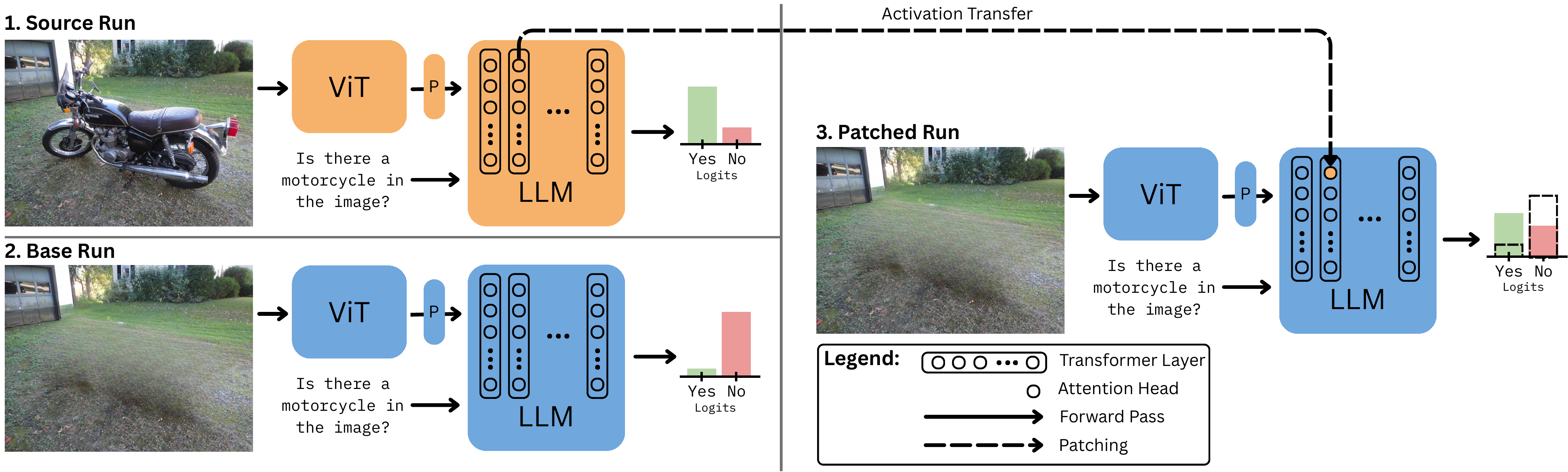}
    \caption{\textbf{Causal mediation via activation patching.} We compare three model runs: (1) the source run, where the object is present and the model produces the correct answer; (2) the base run, where the object is removed and the model fails; and (3) the patched run, where we transfer hidden activations from a selected attention head in the source run into the base run. Improvements in the patched prediction indicate that the transferred head carries task-relevant information. This procedure is repeated head-wise to quantify each head’s causal contribution.}
    \label{fig:apx:cma_overview}
\end{figure}

\vspace{-0.5em}
\paragraph{Method.} We apply CMA using activation patching to identify which attention heads causally contribute to solving the visual task. 
For each example, we construct two versions of the \textit{same} image: a source image, where the relevant object is present, and a base image, where the object has been removed using a diffusion-based inpainting model (see \cref{subsec:dataset}). We use 50 images from this curated subset.
Separately for each attention head, we extract the hidden activations from the source run and patch them into the forward pass of the base run, yielding the counterfactual output $y^{\star}$. As in our attention-blocking experiments (\cref{sec:attn_knockout}), we patch all activations associated with the prompt tokens. An overview of the setup is shown in \Cref{fig:apx:cma_overview}.

\noindent All outputs are evaluated under teacher-forcing using token-level perplexity. 
Since fixed template tokens (e.g., brackets in a bounding-box answer) are predicted with near-deterministic probability and would artificially dominate perplexity, we mask these tokens out when computing the score. 
Let $P_{\text{base}}$, $P_{\text{src}}$, and $P_{\text{patched}}$ denote the perplexity of the base, source, and patched runs, respectively. We quantify the causal contribution of a component using the Mediation Fraction (MF):
\begin{align}
    \text{MF} = \frac{P_{\text{base}} - P_{\text{patched}}}{P_{\text{base}} - P_{\text{src}}}.
\end{align}
MF measures how much of the performance gap between base and source is closed by the patched component: an MF close to 1 indicates that the patched head fully mediates the task-relevant information; $\text{MF} \approx 0$ implies no contribution; and $\text{MF} < 0$ indicates misleading or interfering effects. 

\noindent For classification, we use the binary query (see \cref{sec:apx:prompts}). The list formulation generates object names in an arbitrary order, which spreads probability mass across many tokens. This leads to per-class perplexity differences that are too noisy for reliable mediation estimates.
In contrast, the binary query produces a single decisive token, which makes it well suited for activation patching. Although it shows a higher hallucination rate in open evaluation, this is not a problem in our setting. The paired source–base design ensures that changes in perplexity reflect causal effects.

\begin{figure}[ht]
    \centering
    \includegraphics[width=\linewidth]{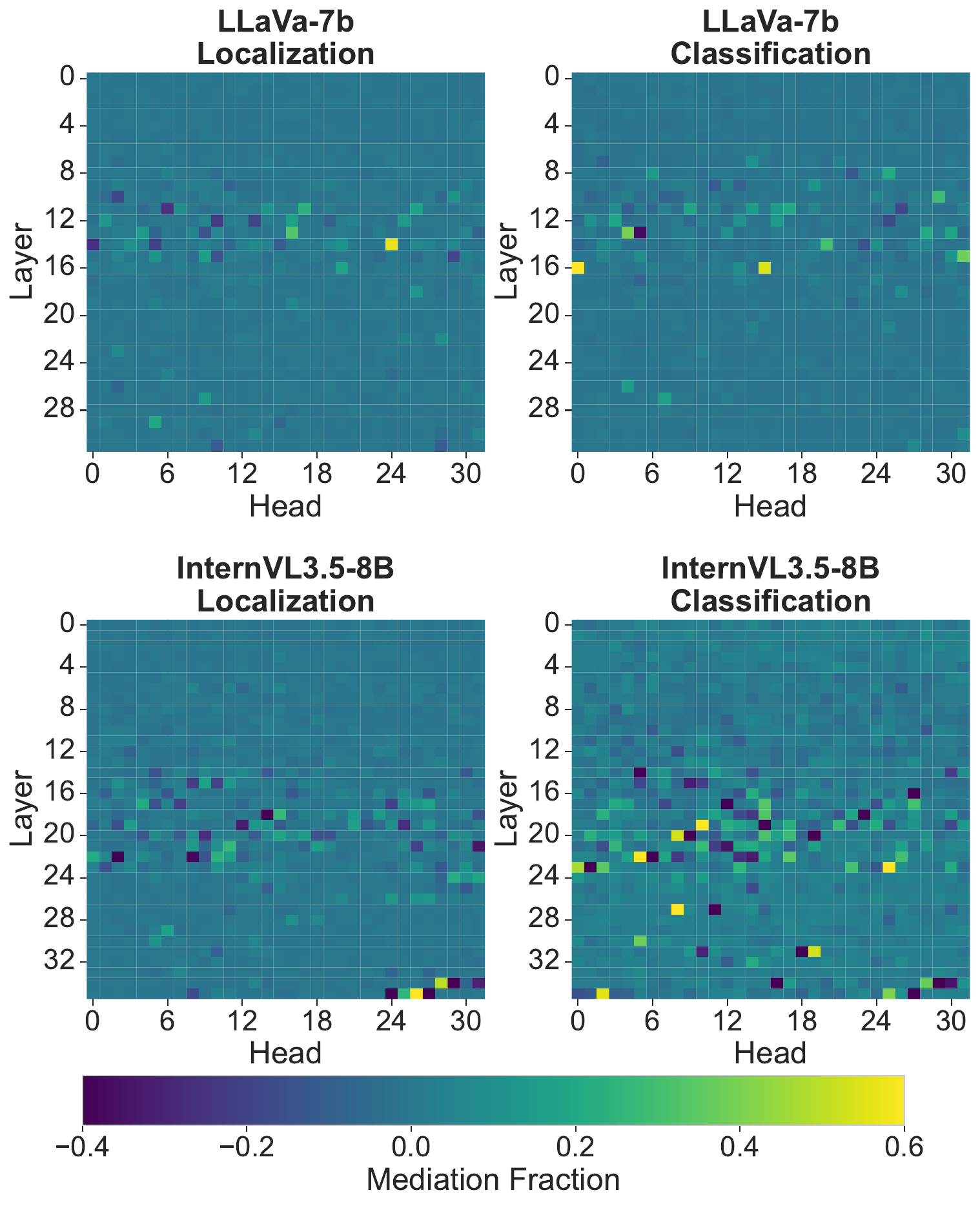}
    \caption{Mediation Fraction (MF) scores for every attention head across all layers, shown separately for the localization and classification tasks. For LLaVA-13B results, refer to appendix \Cref{fig:apx:cma_llava13}.}
    \label{fig:cma_main}
    \vspace{-1em}
\end{figure}

\vspace{-0.5em}
\paragraph{Results.} \Cref{fig:cma_main} reports the Mediation Fraction (MF) for every attention head across all layers. Consistent with the attention-blocking experiment (\cref{sec:attn_knockout}), the causal mediation analysis locates the core processing region in the early–mid layers for LLaVA and in the mid–late layers for InternVL. For LLaVA, most non-zero MF values cluster around layers 11–16 for both localization and classification, whereas for InternVL the dominant heads appear in layers 16–22. Outside these ranges, nearly all heads exhibit MF scores close to zero, indicating that the vast majority of attention heads are not causally involved in the task. \\\\
\noindent In addition, the distribution of MF scores across heads is highly sparse for both tasks. Only a small number of attention heads exhibit large mediation values, indicating that the bulk of task-relevant information is carried by a compact set of specialized heads. The vast majority of heads contribute negligibly, with MF values near zero, suggesting that they neither facilitate nor interfere with the task. The two tasks differ in how this information is distributed: localization relies on a more concentrated subset of heads, whereas classification engages a broader set of heads dispersed across earlier and intermediate layers. Despite these differences, we do observe a limited number of shared heads. Among the top-10 heads ranked by MF, only two heads in LLaVA and one in InternVL are shared across both tasks; even when expanding to the top-50 heads, the overlap remains limited: 20 for LLaVA-7B, 15 for LLaVA-13B, and 18 for InternVL. This pattern indicates that while some early computations are reused, the dominant mediators of localization and classification are largely distinct. \\
\noindent Finally, we observe a substantial number of negative MF values, especially in InternVL, suggesting that certain activations from the source image are incompatible with those of the base image. In these cases, patching increases perplexity, indicating that these heads encode counter-evidence or contextual signals that conflict with the ground-truth output. \\\\
\noindent Overall, these results reveal that VLMs implement visual reasoning through highly selective, low-redundancy pathways: only a handful of attention heads mediate nearly the entire causal effect, and different tasks rely on different subsets of these components, reflecting functional specialization within the model’s internal computation.

\subsubsection{Head Ablation Analysis}
\label{sec:head_ablation}

CMA reveals a small number of attention heads with high MF scores per task. To determine whether these heads are necessary rather than merely correlated with performance, we progressively removed them in a cumulative ablation study.

\vspace{-1em}
\paragraph{Method.}
For each task, we rank all attention heads according to their mean MF score and divide them into two groups: task-critical heads (those with the highest MF scores) and low-importance heads (near zero MF heads). We then perform cumulative ablations on both groups by successively removing a proportion of the total number of heads and measuring the resulting localization accuracy. Ablation is implemented by setting the output of the selected heads to zero during the forward pass. Three settings are evaluated: ablation of localization-critical heads, ablation of classification-critical heads and ablation of low-importance heads as a control baseline.

\begin{figure}[t] 
    \centering
    \includegraphics[width=\linewidth]{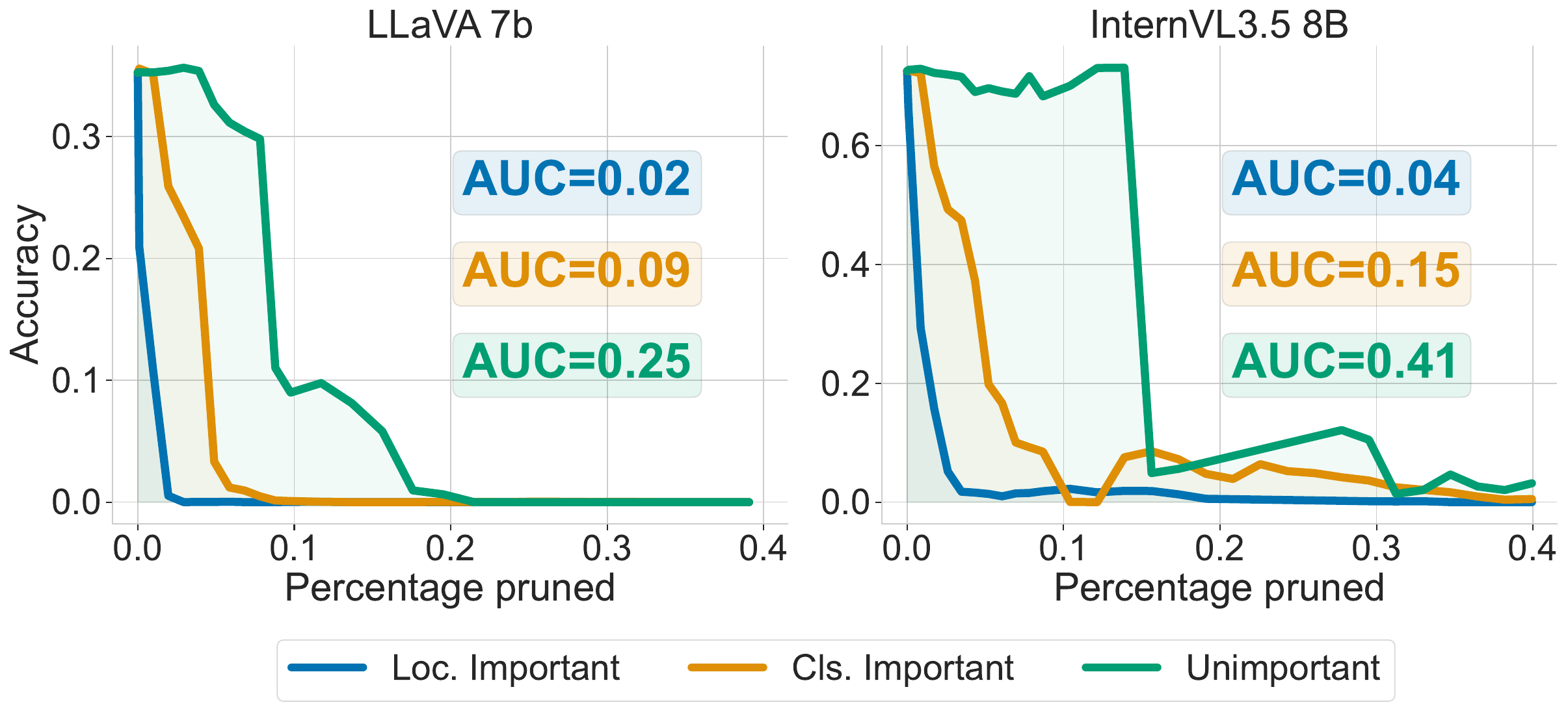}
    \caption{Localization accuracy under cumulative head ablation. Attention heads are ranked by their mean MF and progressively removed. Normalized AUC scores enable method comparison.} 
    \label{fig:head_abl}
    \vspace{-1em}
\end{figure}

\vspace{-1em}
\paragraph{Results.} 
As shown in \Cref{fig:head_abl}, ablating task-critical heads leads to a substantially larger drop in localization accuracy than removing an equal number of low-importance heads. This confirms that the CMA ranking identifies components that are causally necessary for the task rather than simply being correlated with performance. In contrast, removing unimportant heads results in only a slight decline in performance, even when a large fraction of them is removed, indicating that task-relevant information is concentrated in a small subset of attention heads.
Despite the limited overlap between classification-critical and localization-critical heads, ablating classification-critical heads still causes a strong degradation in localization accuracy. This indicates that localization depends on intermediate object-identification representations rather than operating independently.
Together with the attention-knockout results (\cref{sec:apx:attn_blk}), these findings provide causal evidence for a sequential processing mechanism in which the model first identifies the object and subsequently determines its spatial extent using a smaller set of specialized heads.

\section{Related Work}
Prior mechanistic interpretability studies of VLMs primarily examine high-level reasoning, VQA behavior, or hallucinations \cite{basu2024understanding,palit2023towards,jiang2024interpreting,kaduri2024_vision_of_vlms}. \citet{li2026causal} traces object representations across layers and modalities to mitigate hallucinations, while \citet{yu2025how} uses layer-wise probing to reveal a stage-wise processing hierarchy in VLMs. We build on these approaches but focus specifically on where and how localization information is extracted and transformed, using causal mediation analysis at the attention-head level.

\noindent A second thread of research documents systematic failure modes across a range of VLM capabilities, including single-object classification \cite{zhang2024why}, multi-object classification \cite{campbell2024understanding}, counting \cite{rane2024can}, visual search \cite{campbell2024understanding}, and spatial reasoning \cite{chen2025why}. These studies reveal significant gaps in how VLMs ground semantic and spatial information, but they do not examine object localization or identify where in the model spatial representations emerge. 

\noindent In parallel, several recent works aim to improve grounding and localization capabilities in VLMs through architectural changes, additional training strategies, or specialized datasets \cite{pantazopoulos2025towards,zhang2024llava,peng2024grounding,deitke2025molmo,bai2023qwenvlversatilevisionlanguagemodel}. While these models demonstrate enhanced performance on grounding benchmarks, their internal computational mechanisms remain largely unexplored. 
\section{Discussion and Limitations}
Our findings shed light on the fundamental mechanisms through which VLMs capture and encode spatial structure.
Our experiments reveal that positional information is reconstructed in the LLM, rather than relying on positional information encoded in the visual backbone.
Localization depends on a containerization process in which object tokens collectively define spatial boundaries, while the internal spatial or semantic arrangement of these tokens plays only a minor role.
Moreover, causal mediation analysis shows that only a very small number of attention heads are responsible for both tasks, emerging in the early–mid layers for LLaVA and in the mid–late layers for InternVL, with classification and localization relying on largely distinct sets of specialized heads.
Cumulative head ablation further reveals that localization causally depends on classification-critical heads, providing evidence consistent with a sequential processing mechanism in which the model first identifies the object and subsequently determines its spatial extent.
These results refine our understanding of VLMs and open directions, including targeted head fine-tuning or grounding-aware attention supervision.

\noindent 
Our study uses a filtered COCO subset where images contain multiple objects but only one per queried category. This allows us to isolate foundational spatial mechanisms, and the same approach can be naturally extended to more complex images and tasks. 
We apply CMA to attention heads and analyze fixed models, leaving other components and training dynamics unexplored. Extending this framework to segmentation, relational grounding, video, and additional architectures is a promising direction for future work.
\section*{Acknowledgments}
 This work was funded by the Deutsche Forschungsgemeinschaft: DFG project 5368 (``Abstract REpresentations in Neural Architectures (ARENA)'') and DFG project 539642788, RO 6458/5-1 (``Learning from the Environment Through the Eyes of Children (LEECHI)''). We gratefully acknowledge support from The Hessian Center For Artificial Intelligence and Goethe-University (NHR Center NHR@SW) for providing the computing and data-processing resources needed for this work.

{
    \small
    \bibliographystyle{ieeenat_fullname}
    \bibliography{main}
}

\clearpage
\onecolumn
\setcounter{page}{1}
\newpage
\begin{center}
    \Large
    \textbf{\thetitle}\\
    \vspace{0.5em}Supplementary Material \\
    \vspace{1.0em}
\end{center}

\section{Dataset}
We provide additional details on dataset construction, task prompting, and representative examples.

\subsection{Dataset Filtering Details} \label{sec:apx:data_filtering}

We evaluate on the COCO validation split~\cite{lin2014microsoft} with label corrections from~\cite{cocorem}, and apply the following filtering steps to improve annotation quality and satisfy the requirements of our experimental setup.

\begin{enumerate}[leftmargin=1cm,itemsep=4pt, topsep=4pt]
    \item \textbf{Object size.} We discard objects occupying less than $0.4\%$ or more than $60\%$ of the image area. Extremely small objects provide insufficient visual detail, while very large objects dominate the field of view.

    \item \textbf{Resolution.} Images with a minimum side length below $200$ px are removed to ensure sufficient spatial resolution.

    \item \textbf{Uniqueness filtering.} We retain only images that contain a single instance of the queried object category in order to avoid multiple valid answers for the same query.
\end{enumerate}

\subsection{Prompts and Input Template}
\label{sec:apx:prompts}

For each image--object pair, we query the model either to localize a target object or to classify the objects present in the image.

\paragraph{Input template.}
All models receive inputs in the following format:
\[
\texttt{\{System\} \{Image\} \{Task\}}
\]
\noindent where \texttt{System} is the model-specific system prompt prepended to every query, \texttt{Image} refers to the embedded visual tokens, and \texttt{Task} is the task-specific instruction described below. Given this combined input, the model generates its response autoregressively.

\paragraph{Localization.}
For object localization, the model is prompted to predict the bounding box of a given target class:

\begin{quote}
\texttt{Please provide the bounding box coordinates of the \{class\}.}
\end{quote}

\noindent where \texttt{\{class\}} is the name of the target object.

\paragraph{Classification.}
For object classification, we instruct the model to enumerate all objects present in the image while restricting the answer space to the predefined category set:

\begin{quote}
\texttt{List all objects in the image. Choose only from \{class$_1$\}, \{class$_2$\}, \ldots}
\end{quote}

\noindent This formulation avoids the need to handle a large variety of free-form answers, which is particularly problematic for COCO due to its broad and diverse category set.

\paragraph{Alternative binary formulation for classification.}
A natural alternative is to pose a binary query per category:

\begin{quote}
\texttt{Is there a \{class\} in the image?}
\end{quote}

\noindent However, this formulation substantially increases the incidence of object hallucinations. To quantify this effect, we measure the false positive rate
\[
\mathrm{FPR} = P(\mathrm{detected=True} \mid \mathrm{has\_object=False})
\]
on the manipulated dataset described in \Cref{subsec:dataset}, where the queried object is absent. Across InternVL3-5 8B and LLaVA-1.5 7B/13B, the binary query yields consistently higher FPRs (0.56--0.70) than the list-based formulation (0.27--0.45). For this reason, all main experiments use the list-based prompt.

\clearpage
\subsection{Image Examples}
\begin{figure}[ht]
    \centering
    \includegraphics[width=\linewidth]{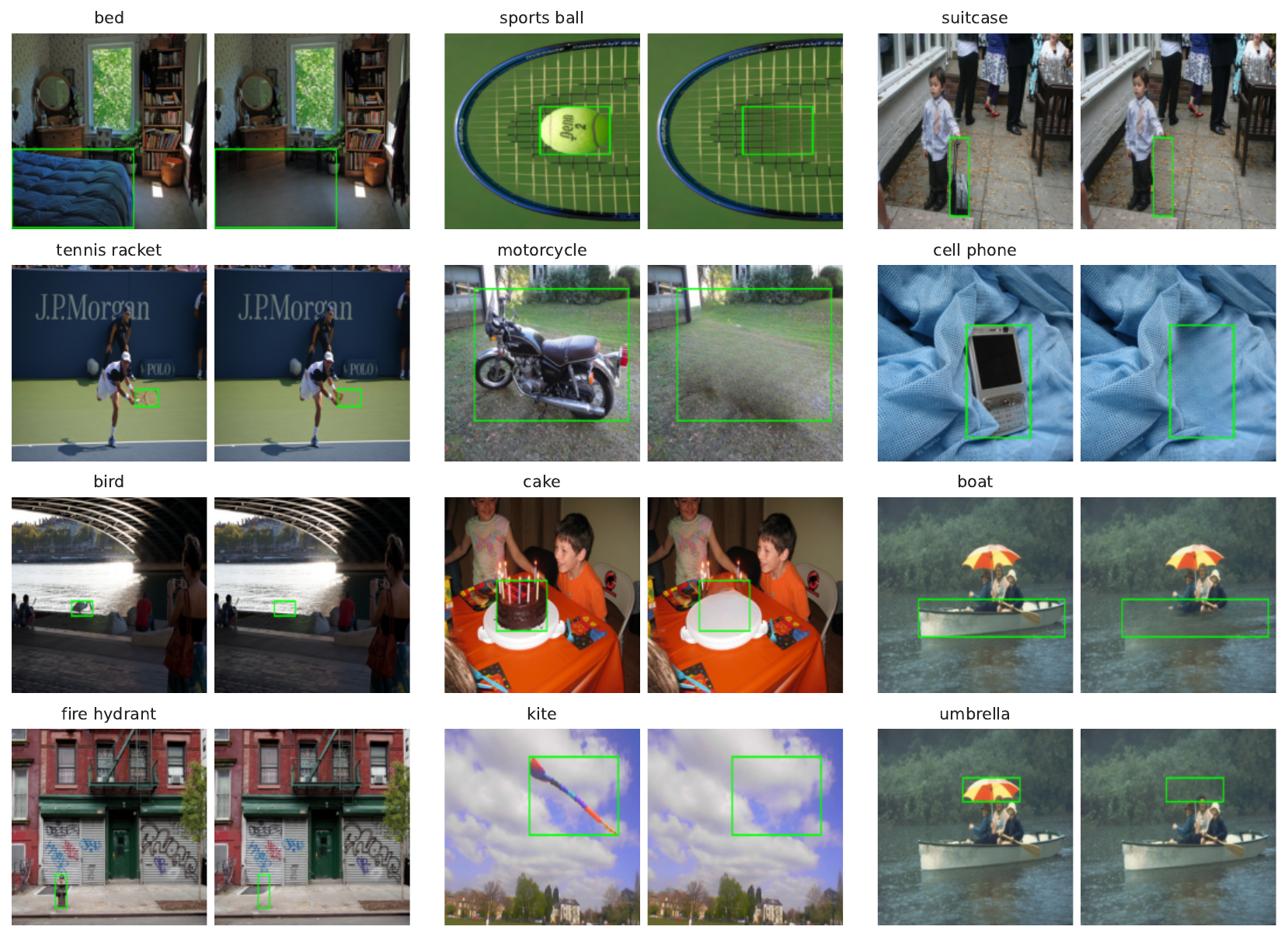}
    \caption{\textbf{Dataset examples.} Example images from the COCO validation set where exactly one object per image is removed (highlighted by its original bounding box) and the background is filled using an inpainting strategy~\cite{lama}. This procedure allows us to filter the dataset for potential hallucinations: if the model can still detect the removed object purely from contextual cues, it undermines the validity of our grounding analysis.}
    \label{fig:apx:object_removal_example}
\end{figure}

\clearpage
\section{Ablation Study}

\subsection{Visualization of Masking}
\begin{figure}[ht]
    \centering
    \includegraphics[width=\textwidth]{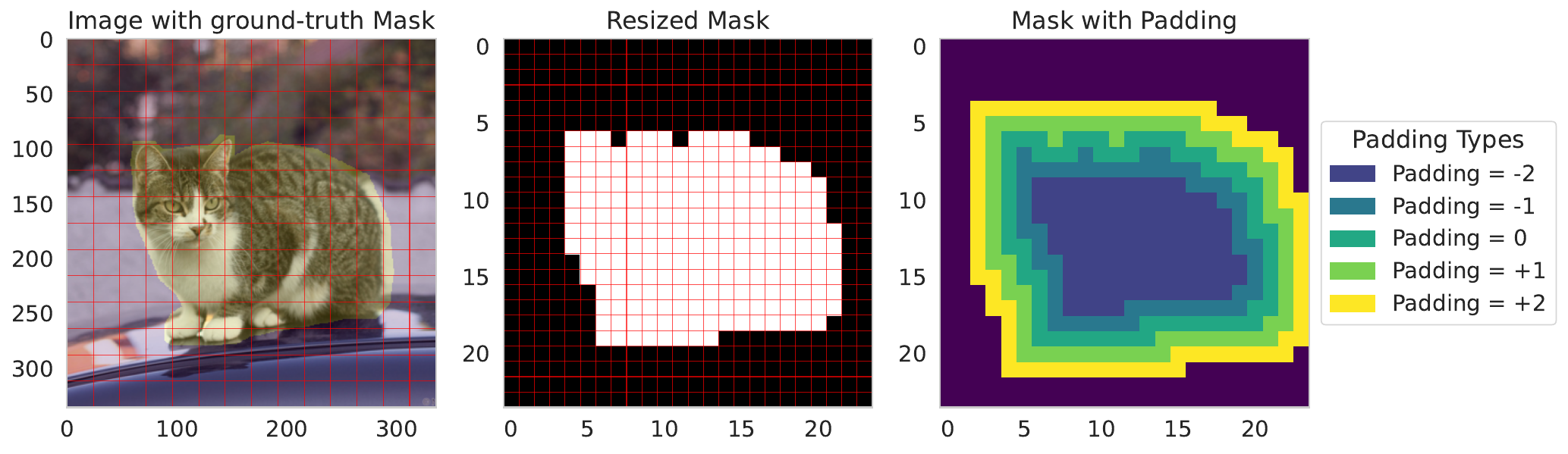}
    \caption{\textbf{Visualization of object mask for the ablation experiment}. Left: the original 336×336 input image in pixel space with an annotated object mask. This mask is mapped onto the 24×24 token grid of the vision transformer, where a token is selected if it has any pixel overlap with the original mask. Right: examples of padding applied to the token mask. Negative padding removes adjacent tokens and shrinks the ablated region, while positive padding adds neighboring tokens and expands it. }
    \label{fig:apx:abl_pad_ex}
\end{figure}

\subsection{Standard Deviations of Random Ablation Experiment}
\begin{table}[ht]
\setlength{\tabcolsep}{2pt}
\sisetup{
  detect-weight=true,
  detect-family=true,
  table-number-alignment = center,
  table-text-alignment = center
}
\centering
\small
\begin{tabular}{
  lrcc|cc|cc
}
\toprule
Models: & \multicolumn{3}{c}{\textbf{LLaVa 7B}} 
& \multicolumn{2}{c}{\textbf{LLaVa 13B}} 
&  \multicolumn{2}{c}{\textbf{InterVL3.5 8B}} \\
\cmidrule(lr){1-1} \cmidrule(lr){2-4} \cmidrule(lr){5-6}\cmidrule(lr){7-8}
\textbf{\shortstack{Ablation\\Strategy}} & 
\multicolumn{1}{c}{\shortstack{Token\\(\%)}} &
 Loc. (\%)&\multicolumn{1}{c|}{Cls. (\%)} &
 Loc. (\%)&\multicolumn{1}{c|}{Cls. (\%)} &
 Loc. (\%) &\multicolumn{1}{c}{Cls. (\%)} \\
\midrule
Baseline               & \textit{0}  & 35.34 & 58.10  & 46.98 & 65.30  & 72.64 & 83.30   \\\midrule
\multirow{7}{*}{\shortstack{Random\\(3 seeds)}} 
 & \textit{1}  & 35.52 $\pm$ 0.21 \up{0.2} & 57.98 $\pm$ 0.18 \down{0.1} & 46.74 $\pm$ 0.08 \down{0.2} & 64.86 $\pm$ 0.40 \down{0.4}  & 72.32 $\pm$ 0.23 \down{0.3} & 83.44 $\pm$ 0.51 \up{0.1} \\
 & \textit{4}  & 35.57 $\pm$ 0.09 \up{0.2} & 57.35 $\pm$ 0.18 \down{0.8} & 45.99 $\pm$ 0.39 \down{1.0} & 64.68 $\pm$ 0.23 \down{0.6}  & 72.30 $\pm$ 0.39 \down{0.3} & 83.33 $\pm$ 0.09 \up{0.0} \\
 & \textit{8}  & 35.09 $\pm$ 0.08 \down{0.2} & 56.58 $\pm$ 0.29 \down{1.5} & 45.25 $\pm$ 0.08 \down{1.7} & 63.89 $\pm$ 0.14 \down{1.4}  & 71.71 $\pm$ 0.21 \down{0.9} & 83.10 $\pm$ 0.35 \down{0.2} \\
 & \textit{16} & 33.71 $\pm$ 0.50 \down{1.6} & 56.39 $\pm$ 0.65 \down{1.7} & 43.76 $\pm$ 0.16 \down{3.2} & 63.92 $\pm$ 0.24 \down{1.4}  & 70.46 $\pm$ 0.20 \down{2.2} & 83.02 $\pm$ 0.39 \down{0.3} \\
 & \textit{24} & 31.92 $\pm$ 0.23 \down{3.4} & 55.72 $\pm$ 0.52 \down{2.4} & 41.95 $\pm$ 0.29 \down{5.0} & 63.51 $\pm$ 0.29 \down{1.8}  & 68.81 $\pm$ 0.42 \down{3.8} & 82.83 $\pm$ 0.47 \down{0.5} \\
 & \textit{32} & 30.43 $\pm$ 0.13 \down{4.9} & 55.40 $\pm$ 0.34 \down{2.7} & 39.88 $\pm$ 0.33 \down{7.1} & 62.54 $\pm$ 0.52 \down{2.8}  & 66.88 $\pm$ 0.47 \down{5.8} & 82.62 $\pm$ 0.56 \down{0.7} \\
 & \textit{48} & 25.65 $\pm$ 0.29 \down{9.7} & 54.80 $\pm$ 0.83 \down{3.3} & 34.44 $\pm$ 0.16 \down{12.5} & 61.34 $\pm$ 0.66 \down{4.0}  & 59.03 $\pm$ 0.39 \down{13.6} & 81.69 $\pm$ 0.33 \down{1.6} \\
\bottomrule
\end{tabular}
\caption{\textbf{Performance after token ablation.} The baseline corresponds to the model without any token ablation and serves as a reference for randomly ablated tokens. This table additionally reports one standard deviations across three random seeds for the random ablation experiments. See \Cref{tab:abl} for the complete results.}
\label{tab:apx:abl_rand_std}
\end{table}

\subsection{Additional Dataset}
We provide additional results on the Pascal VOC dataset \cite{pascalvoc}. We applied the same pre-processing pipeline as desribed in \Cref{subsec:dataset} , which results in \num{1,957} objects across \num{1,585} images. We observe consisent results as describes in \Cref{sec:abl}.

\begin{table}[h]
\setlength{\tabcolsep}{2pt}
\sisetup{
  detect-weight=true,
  detect-family=true,
  table-number-alignment = center,
  table-text-alignment = center
}
\centering
\small
    \begin{tabular}{lcc|cc|cc}
        \toprule
          Models:  &  \multicolumn{2}{c}{\textbf{LLaVA 7B}} &  \multicolumn{2}{c}{\textbf{LLaVA 13B}} &  \multicolumn{2}{c}{\textbf{InternVL 8B}} \\\cmidrule(lr){1-1} \cmidrule(lr){2-3} \cmidrule(lr){4-5} \cmidrule(lr){6-7}
       Task: &  Loc. & Cls. &  Loc. & Cls. &  Loc. &  Cls. \\
        \midrule
        Baseline &  52.41 & 73.41 & 63.68 & 82.97 & 85.78 & 97.06 \\\midrule
        \textbf{Object} & \textbf{9.52} & \textbf{31.37} & \textbf{18.63} & \textbf{54.78} & \textbf{26.35} & \textbf{52.94} \\
        Gradients & 22.63 & 57.23 & 20.83 & 74.39 & 27.00 & 89.34 \\
        Random & 50.41 & 72.96 & 61.08 & 84.52 & 83.76 & 97.30\\
        \bottomrule
    \end{tabular}
    \caption{Token ablation results on the PascalVOC dataset \cite{pascalvoc}. We compare object-token ablation with ablation of an equal number of high-gradient and randomly selected tokens (3 seeds average).}
    \label{tab:reb:ablation_voc}
\end{table}

\FloatBarrier

\subsection{Object Extension Experiment}
\begin{figure}[ht]
    \centering
     \includegraphics[width=0.4\linewidth]{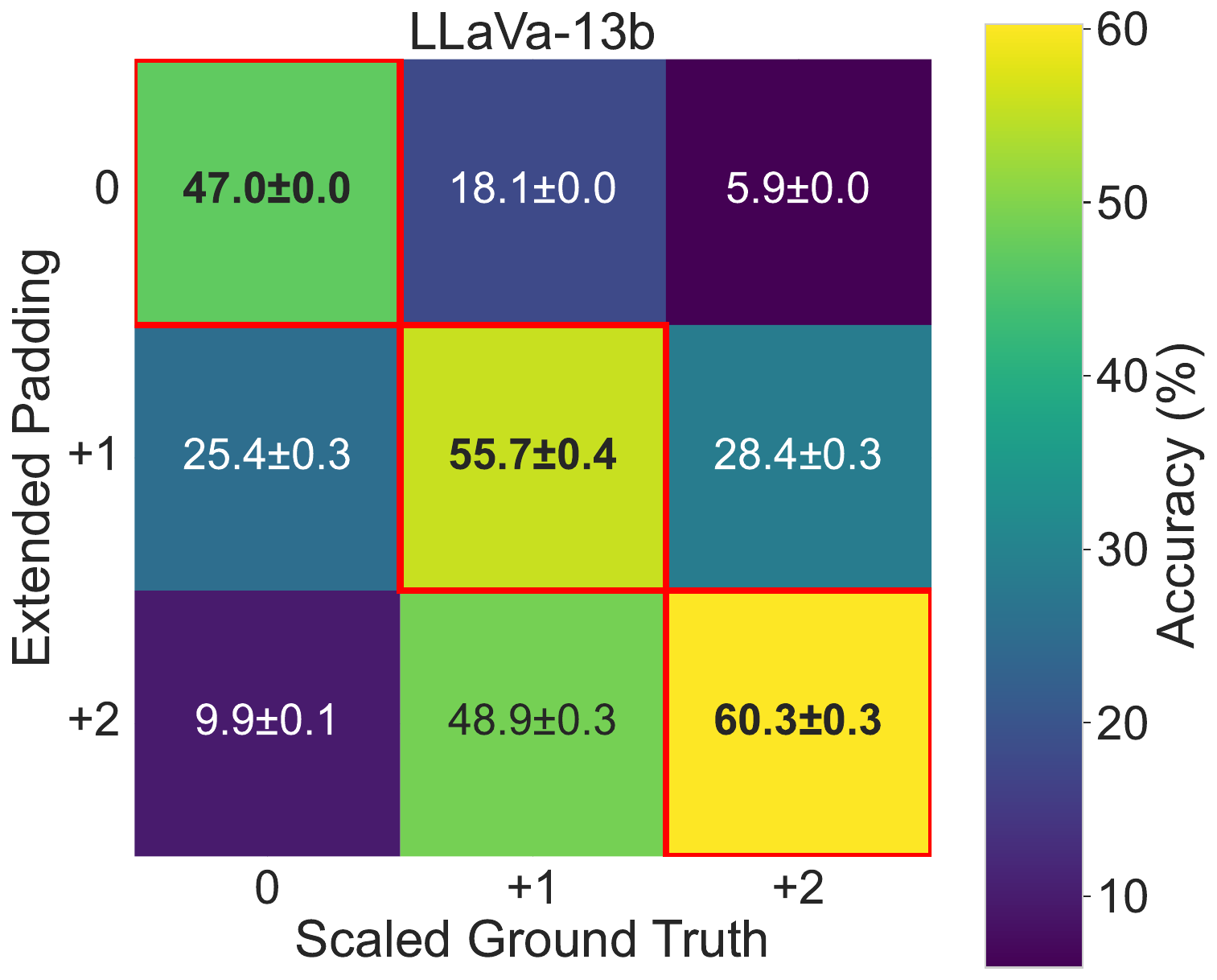}
    \caption{\textbf{Object extension additional Results.} Alignment between predicted and scaled ground-truth bounding boxes under object padding. Each cell shows the mean accuracy between predictions obtained with a given padding level and ground-truth boxes scaled by different amounts. Diagonal entries correspond to matching padding and scaling levels, indicating how well the predicted box size adapts to the artificially enlarged object. Standard deviations are annotated.}
    \label{fig:apx:ablate_add_object_border}
\end{figure}

\begin{figure}[ht]
    \centering
    \includegraphics[width=0.48\linewidth]{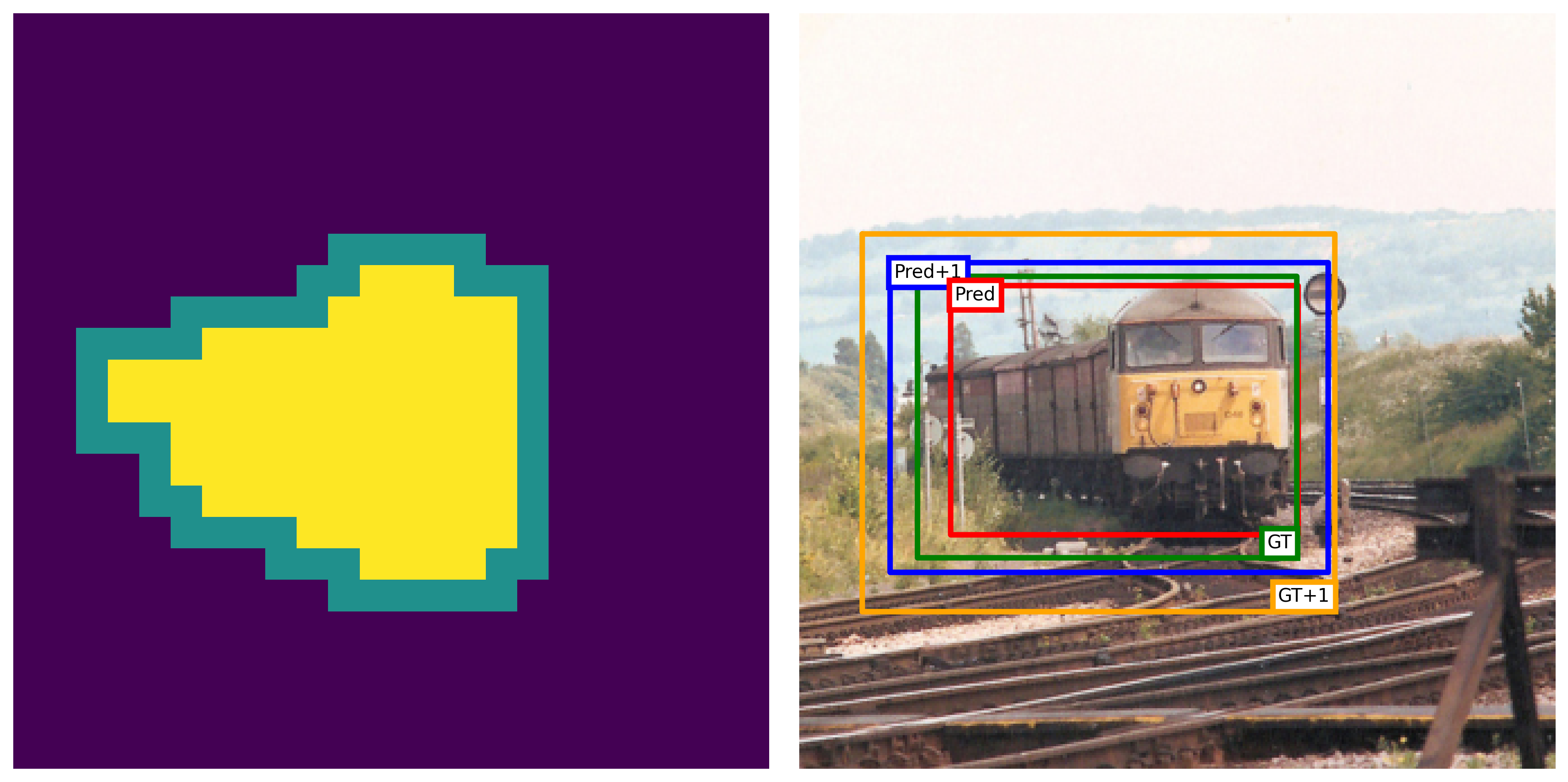}
    \includegraphics[width=0.48\linewidth]{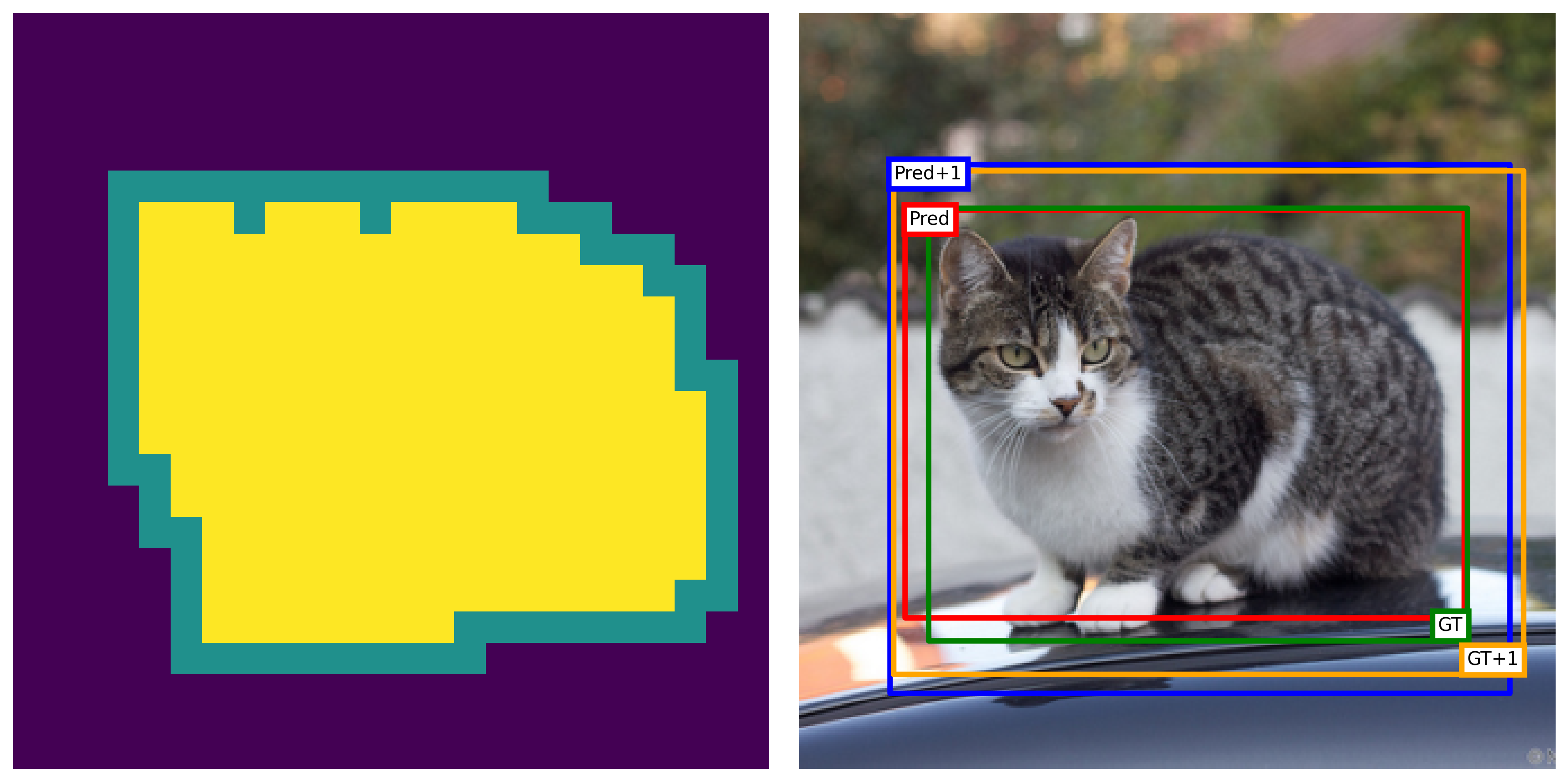}\\
    \includegraphics[width=0.48\linewidth]{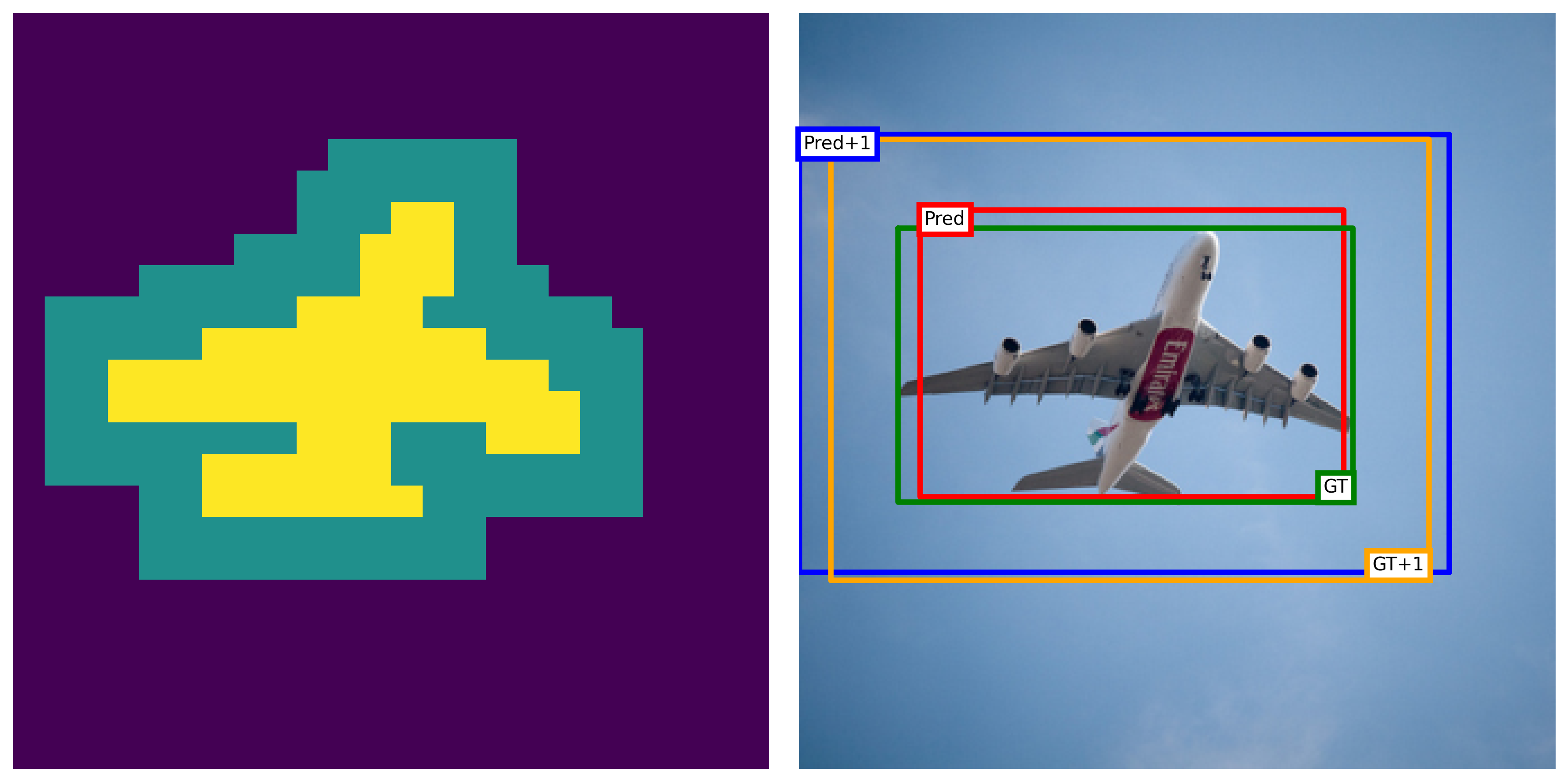}
    \includegraphics[width=0.48\linewidth]{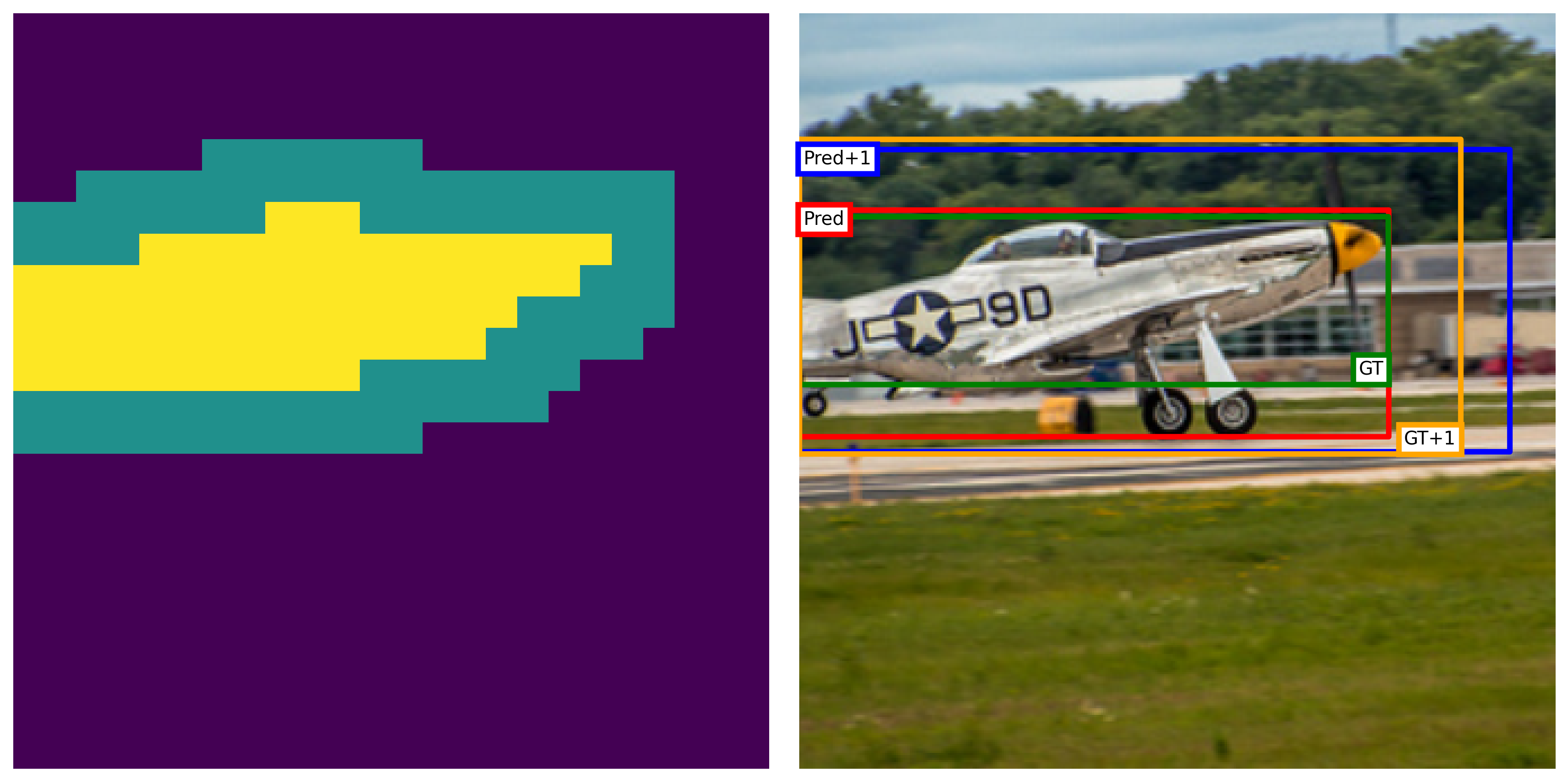}
    \caption{\textbf{Examples of the object extension experiment.} Each image shows the input with its mask in pixel space. The yellow region indicates the original mask, while the green region denotes the padding $p$ added by sampling tokens from the object. The top row corresponds to $p=1$ and the bottom row to $p=2$. We display both the predicted and ground-truth bounding boxes for the original and the extended object. The predicted boxes expand consistently with the mask, suggesting that the model containerizes object tokens to define spatial boundaries.}
    \label{fig:apx:mask_ext}
\end{figure}

\FloatBarrier

\subsection{Ablation Results for Global and Local Views}

\begin{figure}[ht]
    \centering
    \includegraphics[width=0.5\linewidth]{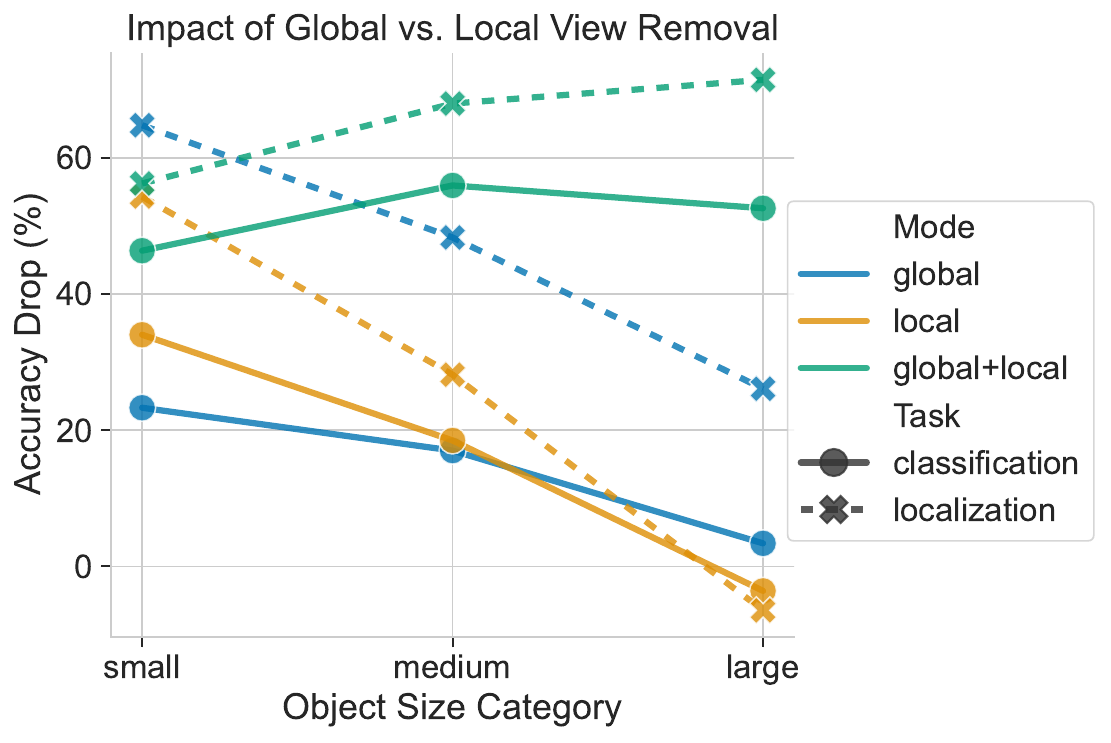}
    \caption{Performance drop for classification and detection when ablating global, local, or both image views across object sizes. We follow the COCO \cite{lin2014microsoft} convention and group targets according to their bounding-box area: \textit{small} ($<32^2$), \textit{medium} ($32^2 \leq \text{area} \leq 96^2$), and \textit{large} ($>96^2$). }
    \label{fig:apx:abl_local_versus_global}
\end{figure}

\clearpage

\section{Positional Information}

\clearpage

\section{Attention Blocking} \label{sec:apx:attn_blk}

\clearpage

\section{Causal Mediation Analysis}
\subsection{Visualization of CMA Method}

\subsection{Additional Results for LLaVa-13b}
\begin{figure}[h]
    \centering
    \includegraphics[width=\linewidth]{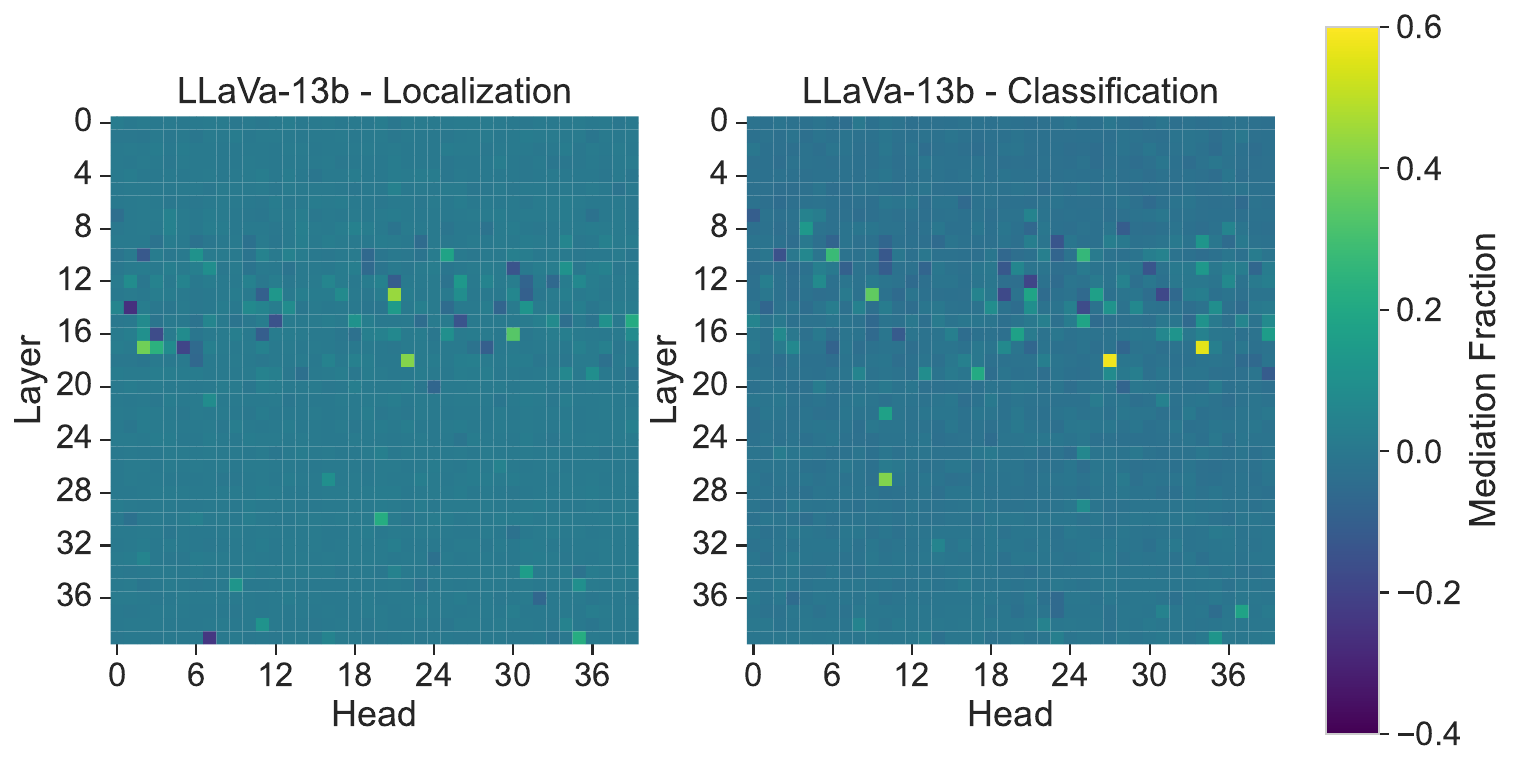}
    \caption{\textbf{Causal Mediation Analysis for LLava13b.} Mediation Fraction scores for every attention head across all layers, shown separately for the detection and classification tasks.}
    \label{fig:apx:cma_llava13}
\end{figure}

\begin{figure}[t] 
    \centering
    \includegraphics[width=0.6\linewidth]{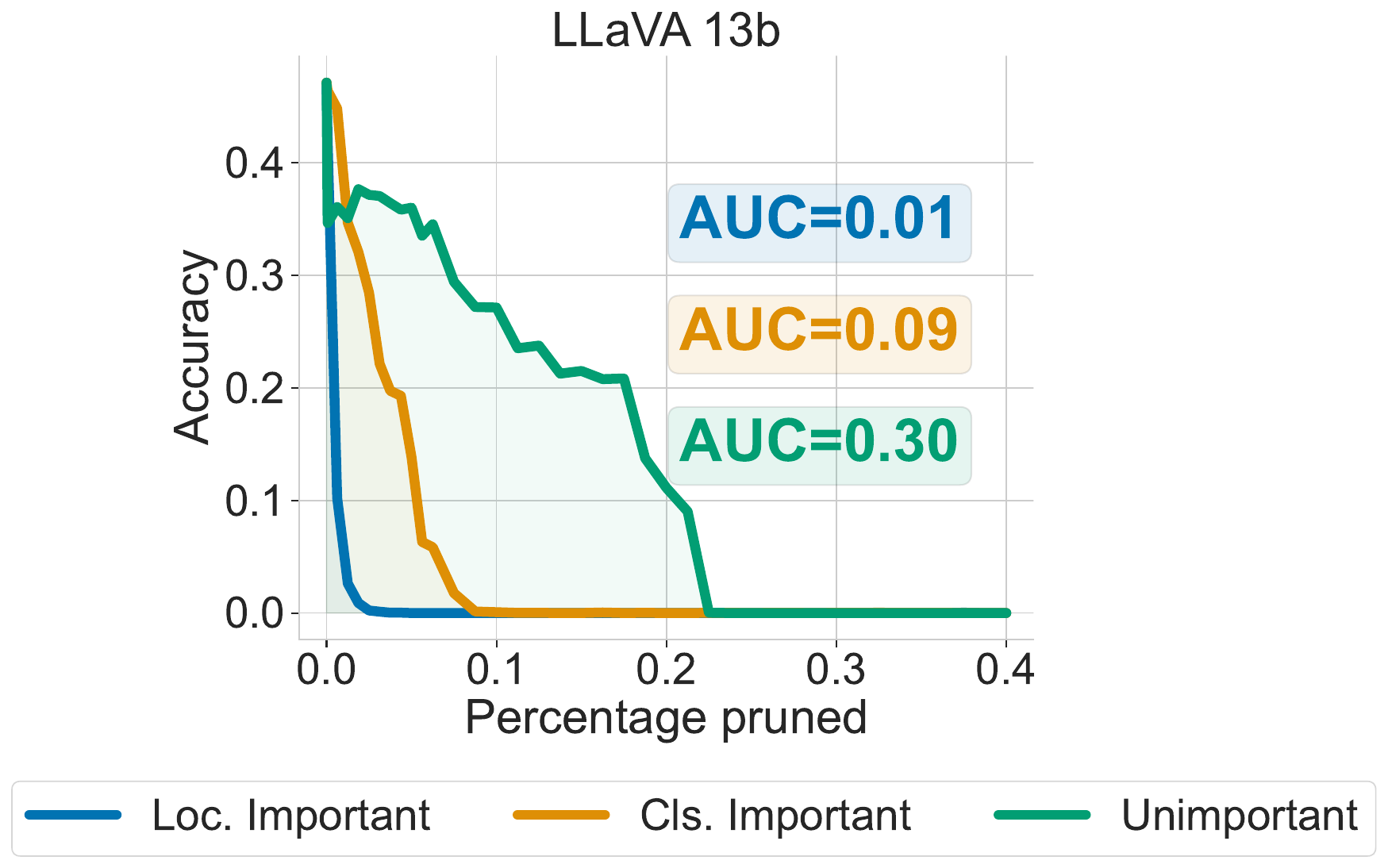}
    \caption{Localization accuracy under cumulative head ablation. Attention heads are ranked by their mean MF and progressively removed. Normalized AUC scores enable method comparison.} 
    \label{fig:apx:head_abl_llava13}
\end{figure}

\end{document}